\documentclass[letterpaper]{article} 
\usepackage{aaai2026}  
\usepackage{times}  
\usepackage{helvet}  
\usepackage{courier}  
\usepackage[hyphens]{url}  
\usepackage{graphicx} 
\usepackage{natbib}  
\usepackage{caption} 
\usepackage{algorithm}
\usepackage{algorithmic}
\usepackage{xspace}
\usepackage{enumitem}
\usepackage[table]{xcolor}
\usepackage{tcolorbox}
\usepackage{booktabs}
\usepackage{multirow} 
\usepackage{subcaption}
\usepackage{amsmath}

\definecolor{teagreen}{HTML}{D0F0C0}
\newcommand{\highlightrow}{\rowcolor{teagreen}}
 
\definecolor{beaublue}{rgb}{0.9, 0.98, 0.85} 
\definecolor{blackish}{rgb}{0.2, 0.2, 0.2}

\definecolor{beaublue2}{rgb}{0.84, 0.9, 0.95}
\definecolor{blackish2}{rgb}{0.2, 0.2, 0.2}

\definecolor{myblue}{rgb}{1.0, 1.0, 1.0} 
\usepackage{newfloat}
\usepackage{listings}
\DeclareCaptionStyle{ruled}{labelfont=normalfont,labelsep=colon,strut=off} 
\floatstyle{ruled}
\newfloat{listing}{tb}{lst}{}
\floatname{listing}{Listing}
\usepackage{amsmath,amsfonts,bm,amsthm}

\def\eqref#1{equation~\ref{#1}}

\def\1{\bm{1}}

\newtheorem{remark}{Remark}

\newcommand{\vphi}{\boldsymbol{\phi}}

\newcommand{\mPi}{{\boldsymbol\Pi}}

\def\vb{{\bm{b}}}

\def\vu{{\bm{u}}}

\def\vx{{\bm{x}}}

\def\vz{{\bm{z}}}

\def\mD{{\bm{D}}}

\def\mM{{\bm{M}}}

\def\mW{{\bm{W}}}
\def\mX{{\bm{X}}}

\def\mZ{{\bm{Z}}}

\def\mPhi{{\bm{\Phi}}}

\DeclareMathAlphabet{\mathsfit}{\encodingdefault}{\sfdefault}{m}{sl}
\SetMathAlphabet{\mathsfit}{bold}{\encodingdefault}{\sfdefault}{bx}{n}

\def\gI{{\mathcal{I}}}

\def\gP{{\mathcal{P}}}

\def\gS{{\mathcal{S}}}

\def\gX{{\mathcal{X}}}

\DeclareMathOperator*{\argmin}{arg\,min}

\newcommand{\abnormal}[1]{{\color[HTML]{E6954B} #1}}  
\newcommand{\normal}[1]{{\color[HTML]{4F7942} #1}}    

\makeatletter
\DeclareRobustCommand\onedot{\futurelet\@let@token\@onedot}
\def\@onedot{\ifx\@let@token.\else.\null\fi\xspace}

\def\eg{\emph{e.g}\onedot}

\def\etc{\emph{etc}\onedot}

\makeatother

\title{Learning Time in Static Classifiers}
\author {
    Xi Ding\textsuperscript{\rm 1}\equalcontrib,
    Lei Wang\textsuperscript{\rm 1, \rm 2}\equalcontrib,
    Piotr Koniusz\textsuperscript{\rm 2, \rm 3, \rm 4, \rm 1}
    Yongsheng Gao\textsuperscript{\rm 1}\thanks{Corresponding author.}
}
\affiliations {
    \textsuperscript{\rm 1}Griffith University, \textsuperscript{\rm 2}Data61/CSIRO, \textsuperscript{\rm 3}University of New South Wales, \textsuperscript{\rm 4}Australian National University\\
    \{x.ding, l.wang4, p.koniusz, yongsheng.gao\}@griffith.edu.au
}
\usepackage{bibentry}

\begin{document}

\maketitle

\begin{abstract}
Real-world visual data rarely presents as isolated, static instances. Instead, it often evolves gradually over time through variations in pose, lighting, object state, or scene context. However, conventional classifiers are typically trained under the assumption of temporal independence, limiting their ability to capture such \textit{dynamics}.
We propose a simple yet effective framework that equips standard feedforward classifiers with temporal reasoning, all without modifying model architectures or introducing recurrent modules. At the heart of our approach is a novel \textit{Support-Exemplar-Query (SEQ) learning paradigm}, which structures training data into temporally coherent trajectories. These trajectories enable the model to learn class-specific temporal prototypes and align prediction sequences via a differentiable soft-DTW loss. A multi-term objective further promotes semantic consistency and temporal smoothness.
By interpreting input sequences as \textit{evolving feature trajectories}, our method introduces a strong temporal inductive bias through loss design alone. This proves highly effective in both static and temporal tasks: it enhances performance on fine-grained and ultra-fine-grained image classification, and delivers precise, temporally consistent predictions in video anomaly detection. Despite its simplicity, our approach bridges static and temporal learning in a modular and data-efficient manner, requiring only a simple classifier on top of pre-extracted features.
\end{abstract}

\begin{links}
    \link{Code}{https://github.com/Darcyddx/time-seq}
\end{links}

\section{Introduction}

Most classification models are trained under the assumption that data points are independent and identically distributed (i.i.d.). However, in many real-world scenarios such as robotics, surveillance, medical imaging, and video analysis, visual data naturally evolves over time \cite{wang2019comparative,wang2023robust,zhu2024advancing,ding2025language}. A person might turn their head, lighting conditions may shift, or an object’s state may gradually change. These temporal variations form coherent, smooth trajectories in feature space. Yet, standard classifiers treat such temporally structured inputs as static, isolated examples, ignoring the rich temporal dynamics inherent in the data.

This mismatch between data reality and training assumptions limits the generalization of conventional classifiers, particularly for tasks requiring robustness to structured perturbations or subtle temporal shifts. While sequence models like RNNs, LSTMs, and Transformers can model temporal information \cite{10123038}, they introduce significant architectural complexity, require temporally annotated data, and are often ill-suited for scenarios with weak or missing frame-level labels \cite{zhu2024advancing}.

In this work, we ask: \textit{Can standard feedforward classifiers reason over time without modifying their architecture, simply through rethinking how we supervise them?} We show the answer is \textit{yes}.
We propose a lightweight, general-purpose training framework that imparts \textit{temporal inductive bias} into static classifiers purely through loss design. Our method operates on smoothly evolving input sequences generated via temporal augmentations that mimic natural transitions such as pose changes or appearance shifts. These sequences pass through a frozen pretrained encoder, followed by a classifier.

At the heart of our framework is a novel \textit{Support-Exemplar-Query (SEQ) learning paradigm} that structures supervision around intra-class temporal patterns. For each query sequence, we align its predictions to class-specific temporal prototypes using a differentiable soft Dynamic Time Warping (soft-DTW) objective. In addition to alignment, we incorporate semantic supervision (via cross-entropy) and a smoothness regularization that penalizes abrupt prediction changes.
This yields a key insight: \textit{temporal reasoning can emerge in static feedforward models purely through supervisory signals, without any architectural modifications or explicit sequence modeling}. Our approach enables such models to learn how class semantics evolve temporally, bridging static and dynamic tasks in a unified, modular, and data-efficient manner.

\begin{figure*}[tbp]
\centering
\begin{subfigure}[t]{0.55\linewidth}
\centering\includegraphics[trim=0cm 0cm 0cm 0cm, clip=true, width=\linewidth]{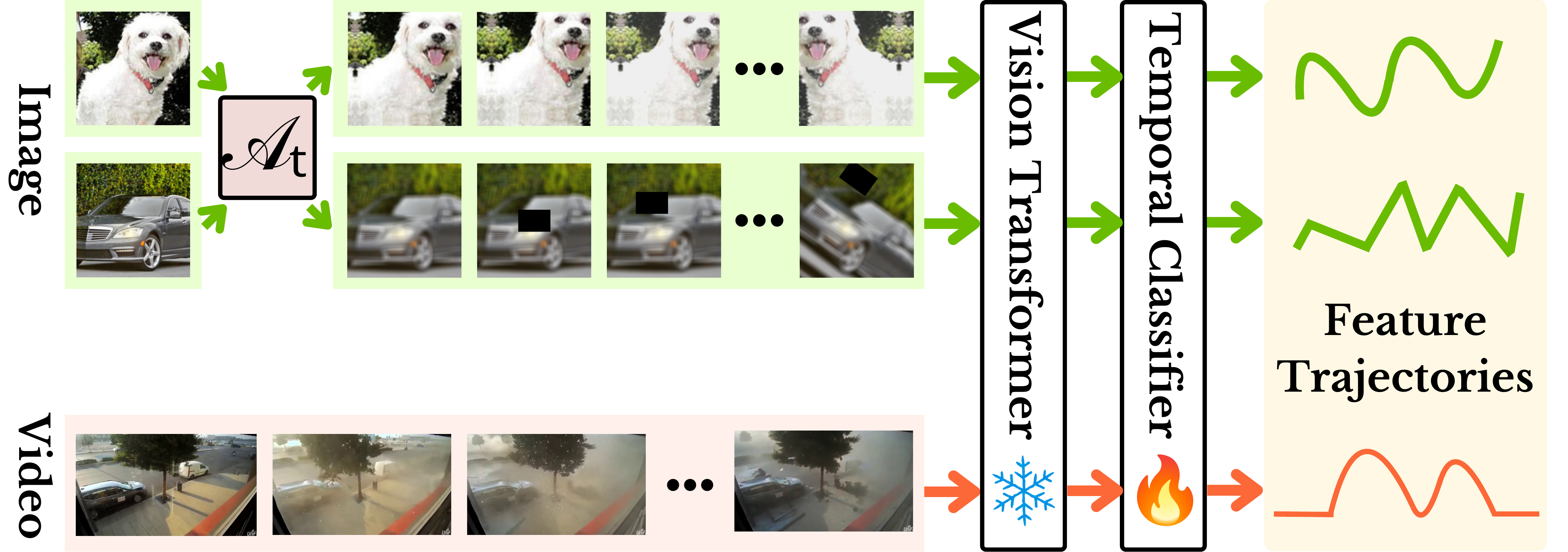} 
\caption{Capturing evolving feature trajectories.}
\label{pipeline1}
\end{subfigure} 
\begin{subfigure}[t]{0.35\linewidth}
\centering\includegraphics[trim=0cm 0cm 0cm 0cm, clip=true, width=\linewidth]{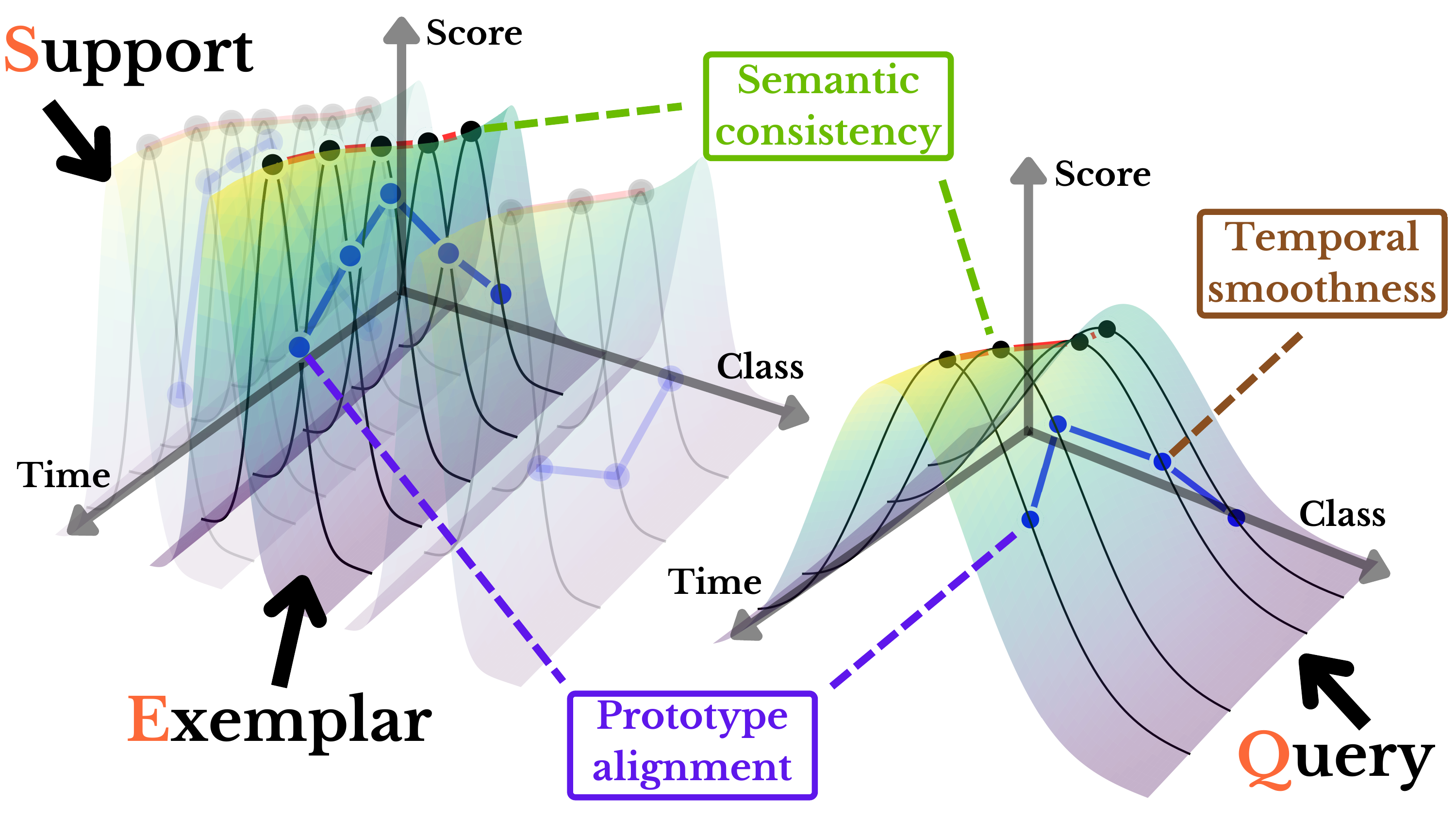}
\caption{The unified multi-term objective.}
\label{pipeline3}
\end{subfigure}
\caption{Overview of our framework. (a) Temporally smooth sequences are generated via time-indexed transformations $\mathcal{A}_t$ (or sourced from natural videos) and processed by a frozen, image-pretrained vision transformer to extract frame-wise features. A lightweight temporal classifier is then trained to produce feature trajectories. (b) These trajectories are optimized using a multi-term objective with the Support-Exemplar-Query (SEQ) learning framework (see Fig. \ref{fig:seq}) to (i) align with class-specific prototype trajectories that capture typical temporal patterns (violet block), (ii) achieve accurate classification through semantic supervision (vivid green block), and (iii) ensure smooth and consistent temporal evolution (gray brown block).}
\label{fig:pipeline}
\end{figure*}

We validate our method on two challenging domains: (i) fine-grained and ultra-fine-grained visual recognition under structured augmentations, where modeling temporal consistency improves generalization, and (ii) frame-level video anomaly detection, where capturing normal temporal behavior enables early and accurate anomaly detection.
Our main \textbf{contributions} are summarized as follows:
\renewcommand{\labelenumi}{\roman{enumi}.}
\begin{enumerate}[leftmargin=0.5cm]
    \item We introduce \textit{SEQ learning}, a novel and effective training paradigm that enables static feedforward classifiers to capture and use temporal class-specific prototype trajectories, without requiring any architectural changes. This \textit{challenges the common assumption} that temporal reasoning requires specialized sequence models. 
    \item We develop \textit{a unified, principled objective} combining soft-DTW temporal alignment, semantic supervision, and smoothness regularization. This framework endows standard classifiers with robust temporal reasoning capabilities purely through loss design and is, to our knowledge, \textit{the first to do so}.
    \item We validate our method on \textit{diverse and challenging tasks}, including fine-grained and ultra-fine-grained image recognition as well as video anomaly detection. Our approach shows significant improvements in generalization, temporal consistency, and anomaly sensitivity while using \textit{only} feedforward architectures.
\end{enumerate}

\section{Related Work}


\noindent\textbf{Temporal modeling in classification.}  
Classical approaches for temporal data rely on architectures explicitly designed to capture sequential dependencies, such as recurrent neural networks (RNNs) including LSTMs and GRUs \cite{10.1162/neco.1997.9.8.1735}, and more recently, attention-based models like Transformers \cite{vaswani2017attention, bertasius2021space,chen2024motion,raj2025tracknetv4}. These methods excel at modeling time series, video, and other sequential data but often require complex architectures, high computational costs, and dense temporal supervision. Their performance degrades when frame-level labels are scarce or when temporal ordering is weak or noisy.

In contrast, our method injects temporal inductive bias directly into the training objective of standard static classifiers, without architectural modifications or recurrent components. By aligning prediction sequences to learned temporal prototypes via soft-DTW, we enable temporal reasoning within simple feedforward models. This approach reduces complexity and broadens applicability to settings where temporal labels or models are unavailable.

\noindent\textbf{Prototype-based learning.}  
Prototype-based classification methods, central to few-shot and metric learning, represent classes by exemplars or centroids in feature space, facilitating generalization from limited data \cite{snell2017prototypical, sung2018learning, wang2022uncertainty}. Extensions to temporal tasks typically learn prototypes with recurrent or convolutional temporal encoders \cite{liu2020prototype, wang2022temporal}.

Our work introduces a novel perspective by defining prototypes in the prediction space as class-specific softmax trajectories over time. Instead of embedding-level comparisons, we align entire prediction sequences to these temporal prototypes using soft-DTW, enforcing not only correct classification but also coherent temporal evolution of predictions. This shift enables temporal supervision even when only static labels are available, representing a significant departure from prior prototype-based methods.

\noindent\textbf{Temporal and smooth augmentations.}  
Data augmentation techniques improve robustness by exposing models to controlled input variations \cite{cubuk2020randaugment, hendrycks2019augmix}. Temporal smoothness regularization and augmentations that mimic natural transitions have been used in video and self-supervised learning to encourage continuity and consistency \cite{sermanet2018time, qian2021spatiotemporal, schiappa2023self, chen2024motion}.

We build upon these ideas by using smooth, structured augmentations to synthesize temporal sequences from static inputs, simulating natural feature trajectories such as pose shifts or illumination changes. Crucially, we use these augmentations not only as regularizers but as core supervisory signals through alignment with temporal prototypes. This enables temporal inductive bias injection even in the absence of real temporal data or frame-level labels.

\noindent\textbf{Learning paradigms for temporal and metric learning.}  
Few-shot and metric learning methods often rely on episodic training paradigms organizing data into support and query sets, promoting generalization from limited exemplars \cite{snell2017prototypical, sung2018learning}. Some approaches extend these paradigms to temporal data by incorporating sequential encoding \cite{liu2020prototype, wang2022temporal,wang2022uncertainty,wang2024meet}.

Our proposed SEQ learning paradigm uniquely structures training data as temporally coherent feature trajectories grouped into support, exemplar, and query roles. This design encourages classifiers to internalize intra-class temporal dynamics through alignment with class-specific prediction prototypes. Unlike existing episodic methods, SEQ integrates soft-DTW alignment on prediction trajectories as a central supervision signal, enabling temporal reasoning without architectural or inference-time complexity. To our knowledge, this is the first framework to combine sequence-level prototype alignment, smooth augmentation-driven trajectory generation, and static feedforward classifiers into a lightweight, unified temporal learning paradigm.

\section{Method}

\begin{figure}[tbp]
    \centering
    \includegraphics[trim=0 0 0 0, clip=true, width=\linewidth]{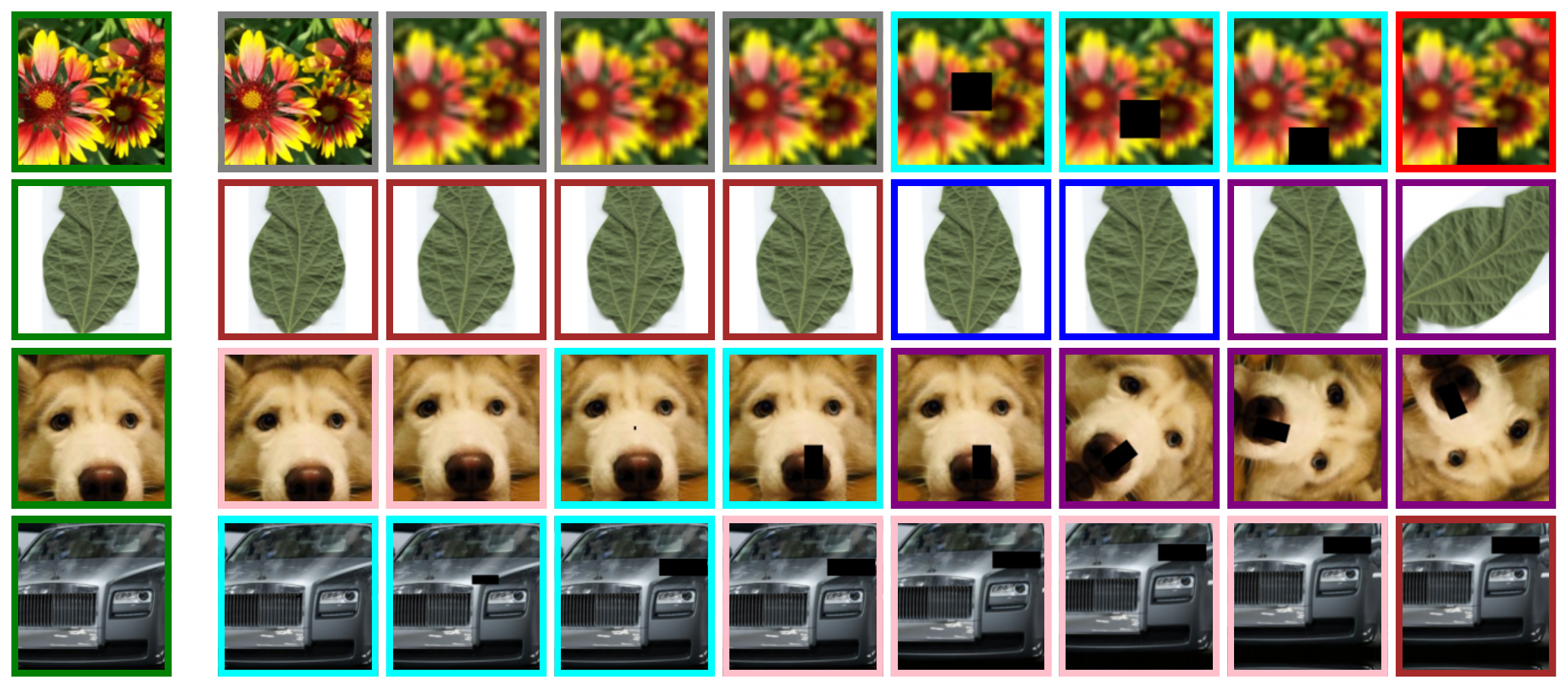}
    \caption{
    Examples from Flowers-102, SoyAging, Stanford Dogs, and Cars show how augmentations create temporal variations from one image. 
    The first column shows originals (green); others apply augmentations by color: flip (red), zoom (blue), rotation (purple), color jitter (orange), shear (brown), translation (pink), blur (gray), and cutout (cyan), enriching the feature space with varied appearances.}
    \label{fig:aug}
\end{figure}

\subsection{Overview}


We introduce a novel framework that infuses temporal inductive bias into static classifiers \emph{without requiring architectural changes or recurrent mechanisms}, see Fig. \ref{fig:pipeline} for framework overview. The central insight of our method is to reinterpret static or sequential inputs as temporally coherent \emph{feature trajectories}, which are then aligned with class-specific \emph{temporal prototypes} using a differentiable sequence alignment procedure. This enables conventional feedforward models to exhibit temporal reasoning capabilities, enhancing their performance in scenarios where temporal consistency is crucial.
It consists of three key components:
\renewcommand{\labelenumi}{\roman{enumi}.}
\begin{enumerate}[leftmargin=0.5cm]
    \item \textbf{Feature trajectories extraction.} We encode static or video inputs into smoothly evolving feature sequences that reflect temporal coherence. 
    \item \textbf{Support-Exemplar-Query (SEQ) learning.} We propose a novel SEQ paradigm that uses intra-class temporal structure by organizing data into support, exemplar, and query trajectories, encouraging the model to learn temporally grounded representations.
    \item \textbf{Multi-term objective.} We optimize a composite loss function comprising (i) \emph{alignment loss} for matching feature trajectories to temporal prototypes, (ii) \emph{semantic supervision} via class labels, and (iii) a \emph{temporal smoothness} regularization to maintain consistency across time.
\end{enumerate}
The result is a robust, temporally aware classifier that generalizes effectively across both synthetic and real-world temporal variations, all while maintaining compatibility with existing architectures. We begin by describing our notation.

\noindent\textbf{Notation.}
Let $\gI_\tau = \{1, 2, \dots, \tau\}$ denote a time index set of length $\tau$. A stacked vector of elements $\alpha_i$ is written as $[\alpha_i]_{i \in \gI_\tau}$, and a matrix formed from elements $\alpha_{ij}$ is denoted $[\alpha_{ij}]_{(i,j) \in \gI_I \times \gI_J}$. Scalars are represented in standard font (\eg, $x$), vectors in bold lowercase (\eg, $\vx$), matrices in bold uppercase (\eg, $\mX$), and tensors in calligraphic font (\eg, $\gX$). The inner product between two matrices $\mPi$ and $\mD$ is defined as the standard Euclidean inner product between their vectorized forms: $\langle \mPi, \mD \rangle \equiv \langle \mathrm{vec}(\mPi), \mathrm{vec}(\mD) \rangle$.

\subsection{Capturing Evolving Feature Trajectories}

\noindent\textbf{Smooth temporal augmentations from images.}  
Static images inherently lack temporal structure, limiting a model’s capacity to learn temporal dynamics or develop temporal reasoning. To overcome this limitation, we synthesize \emph{virtual temporal sequences} from a single image $\gX \in \mathbb{R}^{H \times W \times 3}$ by applying \emph{smooth, time-varying augmentations} over a virtual time index $t \in \gI_\tau$.
Formally, we construct a sequence:
\begin{equation}
\gX = [\gX_1, \gX_2, \ldots, \gX_\tau], \quad \text{with} \quad \gX_t = \mathcal{A}_t(\gX),
\end{equation}
where $\mathcal{A}_t$ denotes a transformation with parameters $\boldsymbol{\theta}_t$ that vary smoothly over time.
Each augmentation parameter $p \in \boldsymbol{\theta}$ (\eg, rotation angle, brightness, translation, \etc) evolves linearly over the sequence length $\tau$:
\begin{equation}
p_t = p_{\text{start}} + \frac{t - 1}{\tau - 1}(p_{\text{end}} - p_{\text{start}}),
\end{equation}
where $p_{\text{start}}$ and $p_{\text{end}}$ are randomly sampled endpoints. This linear interpolation ensures that the transformations evolve continuously across time, mimicking realistic temporal transitions.
The operator $\mathcal{A}_t$ thus combines spatial and photometric effects such as rotation, translation, scaling, brightness, contrast, and blur into a time-indexed transformation:
\begin{equation}
\mathcal{A}_t = \mathcal{T}(\boldsymbol{\theta}_t).
\end{equation}
These augmentations emulate plausible temporal changes, such as gradual pose shifts, or camera zooms, without requiring access to video data or temporal annotations. 
Fig. \ref{fig:aug} shows visualizations of temporal augmentations on images.

\noindent\textbf{Natural temporal sequences from videos.} 
In contrast, video data naturally provides temporal continuity, capturing authentic dynamics such as object motion, scene evolution, and environmental changes. A video clip can be represented as: $\gX = [\gX_1, \gX_2, \dots, \gX_\tau]$,
where each frame $\gX_t$ is temporally correlated with its neighbors, forming a coherent sequence.
This inherent structure encodes rich temporal information that can be directly exploited during training. 

\textbf{Extracting frame-wise features.}  
We adopt a frozen image-pretrained backbone $\mathcal{M}_{\text{Img}}$ to extract frame-wise features from both synthetic and natural sequences. Each frame $\gX_t$ is independently processed:
\begin{equation}
\vz_t = \mathcal{M}_{\text{Img}}(\gX_t),
\end{equation}
yielding a sequence of feature vectors:
\begin{equation}
\mZ = [\vz_1, \vz_2, \dots, \vz_\tau] \in \mathbb{R}^{\tau \times d},
\end{equation}
where $d$ denotes the dimensionality of the extracted features.
We adopt image-pretrained backbones for their rich, transferable visual representations, stability across domains, and efficiency benefits, enabling our classifier to focus solely on learning temporal relationships from strong, frozen features.

We then train a classifier $f$, typically a fully connected layer followed by a softmax activation:
\begin{equation}
\vphi_t = f(\vz_t; \mW), \quad \bm{\Phi} = [\vphi_1, \dots, \vphi_\tau] \in \mathbb{R}^{\tau \times C},
\end{equation}
where $C$ denotes the number of classes. The output $\bm{\Phi}$ is used in classification tasks guided by dedicated loss objectives.
This clean separation between feature extraction and temporal modeling maintains architectural simplicity, while enabling our framework to process both synthetic and real temporal sequences in a unified and scalable manner.

\begin{figure}[tbp]
    \centering
    \includegraphics[trim=0 0 0 0, clip=true, width=0.62\linewidth]{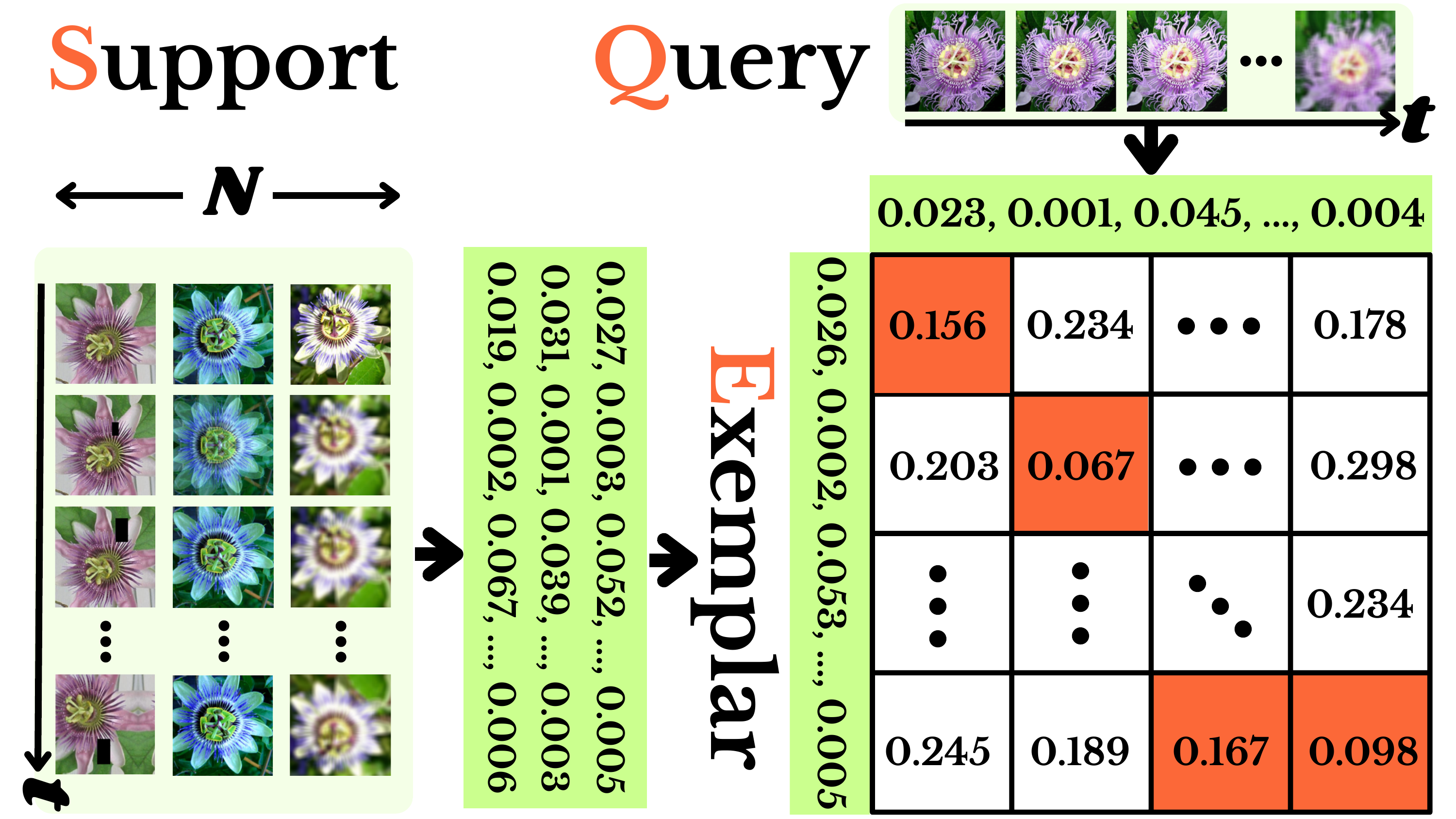}
    \caption{Support-Exemplar-Query (SEQ) models class-consistent temporal dynamics by constructing a \emph{support set} of sequences to form a class-specific \emph{exemplar} that captures typical prediction trajectories over time. A \emph{query sequence} is then aligned against this exemplar to enforce temporal consistency and reveal deviations from expected class behavior.}
    \label{fig:seq}
\end{figure}

\subsection{Support-Exemplar-Query (SEQ) Learning}

We propose \emph{Support-Exemplar-Query (SEQ) learning}, a novel framework for modeling class-consistent temporal dynamics and detecting structural deviations within sequential data. SEQ is built on three key components: (i) a \emph{support set} consisting of class-consistent sequences, (ii) a class-conditioned \emph{exemplar} that summarizes temporal regularities, and (iii) a \emph{query set}, containing sequences evaluated against their corresponding class exemplars for consistency.

The SEQ framework operates in two stages (see Fig. \ref{fig:seq}). First, a \textit{support-query matching} phase selects relevant support sequences for a given query. Second, an \textit{exemplar-query alignment} phase measures the temporal similarity between the query and a synthesized class exemplar via differentiable alignment. The exemplar acts as a dynamic reference that encodes intra-class temporal coherence, facilitating interpretable matching and anomaly detection.

By explicitly capturing the temporal structure within each class and comparing incoming sequences against these learned exemplars, SEQ enables both fine-grained classification and structural deviation detection. Importantly, SEQ uses an \emph{episodic training paradigm}, inspired by few-shot learning, which promotes robust generalization to novel classes and distribution shifts.

\noindent\textbf{Support-query matching.}  
In each training episode, we sample two disjoint subsets from the training data: a \textit{query set} and a \textit{support set}. 
 The query set, denoted as $\gS^{\abnormal{\ast}}$, consists of sequences from various classes (\eg, a batch of training samples), simulating real-world inputs that may be ambiguous or noisy.
%
Given a query sequence $\mPhi^{\abnormal{\ast}} \in \gS^{\abnormal{\ast}}$ with known class label $c$, we construct the corresponding support set $\gS^{\normal{\bullet}} =  \{ \mPhi_n^{\normal{\bullet}} \}_{n \in \gI_N}$ by sampling $N$ additional sequences from the same class $c$. These support sequences are used to synthesize a class exemplar that represents typical temporal score evolution for class $c$.

To compare score sequences of variable lengths, we use the $\gamma$-Soft Dynamic Time Warping (Soft-DTW) distance, a differentiable relaxation of classical DTW. It enables smooth, gradient-based optimization and aggregates alignment costs over multiple plausible warping paths.

Let $\mPhi \!=\! [\vphi_1, \ldots, \vphi_\tau] \!\in\! \mathbb{R}^{\tau \times C}$ and $\mPhi' \!=\! [\vphi'_1, \ldots, \vphi'_{\tau'}] \!\in\! \mathbb{R}^{\tau' \times C}$ denote two sequences of softmax prediction scores. The Soft-DTW distance is computed as:
\begin{equation}
    d^2_{\text{DTW}}(\mPhi, \mPhi') = \text{SoftMin}_{\gamma} \left( \left\{ \langle \boldsymbol{\Pi}, \mD(\mPhi, \mPhi') \rangle \;\middle|\; \boldsymbol{\Pi} \in \gP_{\tau, \tau'} \right\} \right),
\end{equation}
where $\gP_{\tau, \tau'}$ is the set of valid alignment paths between the two sequences, and the alignment cost $\langle \boldsymbol{\Pi}, \mD \rangle$ is computed over the distance matrix $\mD \in \mathbb{R}_+^{\tau \times \tau'}$, defined by:
\begin{equation}
    \mD = \left[ d^2_{\text{base}}(\vphi_m, \vphi'_n) \right]_{(m,n) \in \gI_\tau \times \gI_{\tau'}}.
\end{equation}
Here, $d^2_{\text{base}}(\cdot, \cdot)$ is typically the squared Euclidean distance. 
The SoftMin operator is given by:
\begin{equation}
    \text{SoftMin}_{\gamma}(\boldsymbol{\alpha}) = -\gamma \log \sum_i \exp(-\alpha_i / \gamma),
\end{equation}
where $\gamma \geq 0$ controls the softness of the alignment. As $\gamma \to 0$, it converges to standard DTW; larger $\gamma$ values yield smoother, more flexible alignments.

\noindent\textbf{Query-exemplar alignment.} 
To represent the temporal dynamics of each class, we synthesize an \emph{exemplar} sequence by computing the Fr\'echet mean (or barycenter) of the support set under Soft-DTW. This exemplar captures the average temporal evolution of softmax scores for class $c$, acting as a dynamic prototype for alignment.

Given support set $\gS^{\normal{\bullet}} = \{ \mPhi_n^{\normal{\bullet}} \}_{n \in \gI_N}$ with possibly varying sequence lengths $\tau_n$, the exemplar $\mM^{\normal{\bullet}} \in \mathbb{R}^{\bar{\tau} \times C}$ (where $\bar{\tau}$ is the average length of the sequences in $\gS^{\normal{\bullet}}$) is defined as:
\begin{equation}
    \mM^{\normal{\bullet}} = \argmin_{\mM^{\normal{\bullet}} \in \mathbb{R}^{\bar{\tau} \times C}} \sum_{n=1}^N \frac{w_n}{\tau_n} \, d^2_{\text{DTW}}(\mPhi^{\normal{\bullet}}_n, \mM^{\normal{\bullet}}),
    \label{eq:mean}
\end{equation}
where $w_n \!\in\! \mathbb{R}_+$ are normalized weights satisfying $\sum_{n\!=\!1}^N\! w_n\! = \!1$. This formulation jointly aligns and averages the support sequences, yielding a smooth, representative trajectory of class-consistent score dynamics.

\noindent\textbf{Episodic training paradigm.}  
In each episode, we select a query sequence $\mPhi^{\abnormal{\ast}}$, sample a support set $\gS^{\normal{\bullet}}$ of size $N$, and compute the corresponding class exemplar $\mM^{\normal{\bullet}}$. We then align the query to the exemplar using Soft-DTW, obtaining a class-conditioned similarity score. Note that for generated virtual sequences, we ensure that both the query and its corresponding support sequences undergo identical temporal augmentations. This consistency preserves alignment integrity and allows the model to \textit{focus on class-specific dynamics rather than artificial temporal discrepancies}.
This alignment-based approach equips the model to capture temporal consistency, detect deviations from class patterns, and generalize to new temporal dynamics. Training across diverse episodes encourages abstraction of temporal class structure and adaptability to unseen scenarios.

\subsection{Temporal Classifier with Multi-term Objective}

We propose a lightweight yet expressive classifier that operates over temporal sequences of features using a single fully connected layer followed by softmax. The objective is to train this model to produce temporally coherent, semantically accurate, and class-consistent prediction trajectories. Specifically, the output sequence $\bm{\Phi} = [\bm{\phi}_1, \ldots, \bm{\phi}_\tau]$ should: (i) align with a class-specific prototype trajectory that encodes the typical temporal prediction pattern, (ii) accurately classify each timestep or sequence via semantic supervision, (iii) evolve smoothly over time, avoiding abrupt output changes.
We now present the unified multi-term objective.

\noindent\textbf{Temporal prototype alignment.} 
For each class $c \!\in\! \{1, \!\dots,\! C\}$ in an episode, we construct a class-specific prototype sequence $\mM^{\normal{\bullet}} \!\in\! \mathbb{R}^{\bar{\tau} \times C}$ using the Soft-DTW barycenter method (see~\eqref{eq:mean}). This prototype captures the characteristic temporal evolution of predictions for class $c$.
To align each training query sequence $\bm{\Phi}^{\abnormal{\ast}}$ with its corresponding prototype $\mM^{\normal{\bullet}}$, we minimize their Soft-DTW distance:
\begin{equation}
    \mathcal{L}_{\text{align}} = \frac{1}{|\mathcal{S}^{\abnormal{\ast}}|} \sum_{i=1}^{|\mathcal{S}^{\abnormal{\ast}}|} d^2_{\text{DTW}}\left(\bm{\Phi}_i^{\abnormal{\ast}}, \mM^{\normal{\bullet}}_i\right).
\end{equation}
Here, $\mathcal{S}^{\abnormal{\ast}}$ denotes the set of query sequences, with $|\mathcal{S}^{\abnormal{\ast}}|$ as its cardinality.
This alignment encourages the model to produce temporally structured and class-consistent prediction sequences, even when the input dynamics are nonlinear.

\noindent\textbf{Cross-entropy supervision.} 
To ensure semantic accuracy, we apply standard cross-entropy (CE) loss at either the frame or sequence level, depending on the task:
\begin{enumerate}[leftmargin=0.5cm]
    \item 
    For sequence tasks (\eg, anomaly detection) with frame-level labels $y_t$, cross-entropy is applied at each timestep:
    \begin{equation}
        \mathcal{L}_{\text{CE}} = \frac{1}{\tau |\mathcal{S}^{\abnormal{\ast}}|} \sum_{i=1}^{|\mathcal{S}^{\abnormal{\ast}}|} \sum_{t=1}^\tau \text{CE}(\bm{\phi}_{i,t}^{\abnormal{\ast}}, y_{i,t}).
    \end{equation}

    \item 
    For static tasks (\eg, image classification), predictions are averaged over time to capture visual variations: 
    \begin{equation}
        \mathcal{L}_{\text{CE}} = \frac{1}{|\mathcal{S}^{\abnormal{\ast}}|} \sum_{i=1}^{|\mathcal{S}^{\abnormal{\ast}}|} \text{CE}\left( \frac{1}{\tau} \sum_{t=1}^\tau \bm{\phi}_{i,t}^{\abnormal{\ast}}, y_i \right).
    \end{equation}
\end{enumerate}

This term ensures that predictions convey the correct semantic labels at the \textit{appropriate temporal granularity}.

\noindent\textbf{Temporal smoothness regularization.} 
We add a smoothness loss to enforce temporal stability by penalizing abrupt changes between consecutive predictions:
\begin{equation}
    \mathcal{L}_{\text{smooth}} = \frac{1}{(\tau - 1) |\mathcal{S}^{\abnormal{\ast}}|} \sum_{i=1}^{|\mathcal{S}^{\abnormal{\ast}}|} \sum_{t=2}^{\tau} \left\| \bm{\phi}_{i,t}^{\abnormal{\ast}} - \bm{\phi}_{i,t-1}^{\abnormal{\ast}} \right\|_2^2.
\end{equation}
This encourages the model to produce gradual, interpretable prediction changes that mirror natural dynamics such as motion, progression, or transitions.

\noindent\textbf{Final multi-term objective.}
The complete training loss combines these three components:
\begin{equation}
    \mathcal{L} = \mathcal{L}_{\text{align}} + \alpha \, \mathcal{L}_{\text{CE}} + \beta \, \mathcal{L}_{\text{smooth}},
\end{equation}
where $\alpha$ and $\beta$ are hyperparameters that balance semantic supervision and temporal regularity.
In our experiments, we also incorporate exemplars into both CE loss and temporal smoothness regularization to enhance robustness against semantic variations, perturbations, and class prototype shifts. For sequence tasks, exemplars help capture fine-grained temporal variations within sequences. For image classification, they assist in addressing shifts in class prototypes, thereby improving generalization across diverse conditions.

\begin{table*}[tbp]
\centering
\setlength{\tabcolsep}{0.12em}

\resizebox{\linewidth}{!}{\begin{tabular}{lc|lc|lc|lc}
\toprule
  \multicolumn{2}{c|}{\textbf{Stanford Cars}} & \multicolumn{2}{c|}{ \textbf{Stanford Dogs}} & \multicolumn{2}{c|}{\textbf{Oxford Flowers-102}} & \multicolumn{2}{c}{\textbf{SoyAging}}\\
\cmidrule(lr){1-2} \cmidrule(lr){3-4} \cmidrule(lr){5-6}
\cmidrule(lr){7-8}
Method & Acc & Method & Acc & Method & Acc & Method & Acc \\
\midrule
AP-CNN \cite{9350209} & 95.4 & RAMS-Trans \cite{hu2021rams} & 92.4 & MGE-CNN \cite{9010629} & 95.9 & Cutmix \cite{yun2019cutmix} & 62.3 \\
P2P-Net \cite{yang2022fine} & 95.4 & PMG-V2 \cite{9609669} & 90.7 & SJFT \cite{ge2017borrowing}& 97.0 & DCL \cite{8953746} & 73.2 \\
CP-CNN \cite{9656684} & 95.4 & ViT-NeT \cite{pmlr-v162-kim22g} & 93.6 & OPAM \cite{peng2017object}& 97.1 & ViT \cite{dosovitskiy2021an} & 67.0 \\
TransFG \cite{he2022transfg} & 94.8 & TransFG \cite{he2022transfg} & 92.3 & Cosine \cite{barz2020deep}& 97.2 & DeiT \cite{touvron2021training} & 69.5 \\
ViT-NeT \cite{pmlr-v162-kim22g} & 95.0 & IELT \cite{10042971} & 91.8 & PMA \cite{9103943} & 97.4 & MaskCOV \cite{YU2021108067}& 75.9 \\
DCAL \cite{zhu2022dual} & 95.3 & LGTF \cite{zhu2023learning} & 92.1 & DSTL \cite{cui2018large} & 97.6 & TransFG \cite{he2022transfg}& 72.2 \\
PMG-V2 \cite{9609669} & 95.4 & ACC-ViT \cite{ZHANG2024109979} & 92.9 & MC-Loss \cite{chang2020devil} & 97.7 & SPARE \cite{YU2022108691} & 75.7 \\
GDSMP-Net \cite{KE2023109305} & 95.3 & MP-FGVC \cite{jiang2024delving} & 91.0 & CAP \cite{behera2021context} & 97.7 & Mix-ViT \cite{YU2023109131} & 76.3 \\
MPSA \cite{10638479} & 95.4 & MPSA \cite{10638479} & 95.4 & SR-GNN \cite{bera2022sr} & 97.9 & CLE-ViT \cite{yu2023cle} & {79.0*} \\
\hline
Baseline & 94.7 & Baseline & 93.5 & Baseline & 97.6 & Baseline &  79.6\\
\highlightrow
\textit{w/ feat. traj.} & 95.6 & \textit{w/ feat. traj.} & 96.0 & \textit{w/ feat. traj.} & \textbf{98.4} & \textit{w/ feat. traj.} & 79.8\\
\highlightrow
\textit{w/ feat. traj.\&SEQ} & \textbf{96.1} & \textit{w/ feat. traj.\&SEQ} & \textbf{96.3} & \textit{w/ feat. traj.\&SEQ} & \textbf{98.4} & \textit{w/ feat. traj.\&SEQ} & \textbf{80.0}\\
\bottomrule
\end{tabular}}

\caption{
Performance on fine-grained and ultra-fine-grained recognition datasets.
Baseline and our methods use a single-layer classifier. \textit{w/ feat. traj.} applies smooth temporal augmentations to produce feature trajectories, while \textit{w/ feat. traj.\&SEQ} is our full model with both sequence and temporal modeling. \textbf{Bold} marks the best. Temporal classifier improves performance across image datasets, highlighting the value of temporal cues in fine-grained recognition. * indicates reproduced results.
}
\label{tab:image}
\end{table*}

\begin{figure}[tbp]
\centering
\begin{subfigure}[t]{0.242\linewidth}
\centering\includegraphics[trim=0cm 0cm 0cm 0cm, clip=true, width=\linewidth]{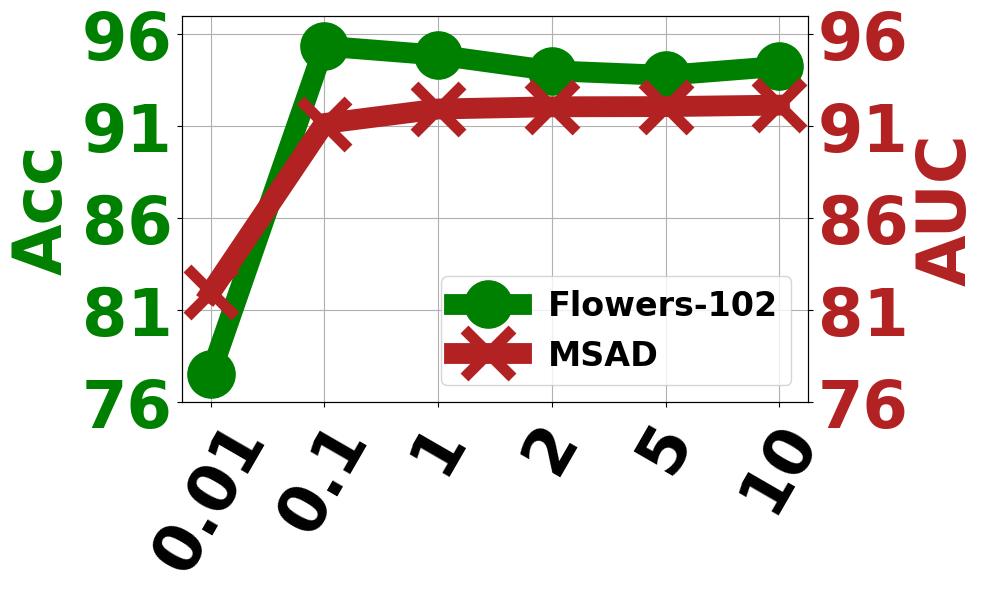}
\caption{$\alpha$}
\label{hyper1}
\end{subfigure}
\begin{subfigure}[t]{0.242\linewidth}
\centering\includegraphics[trim=0cm 0cm 0cm 0cm, clip=true, width=\linewidth]{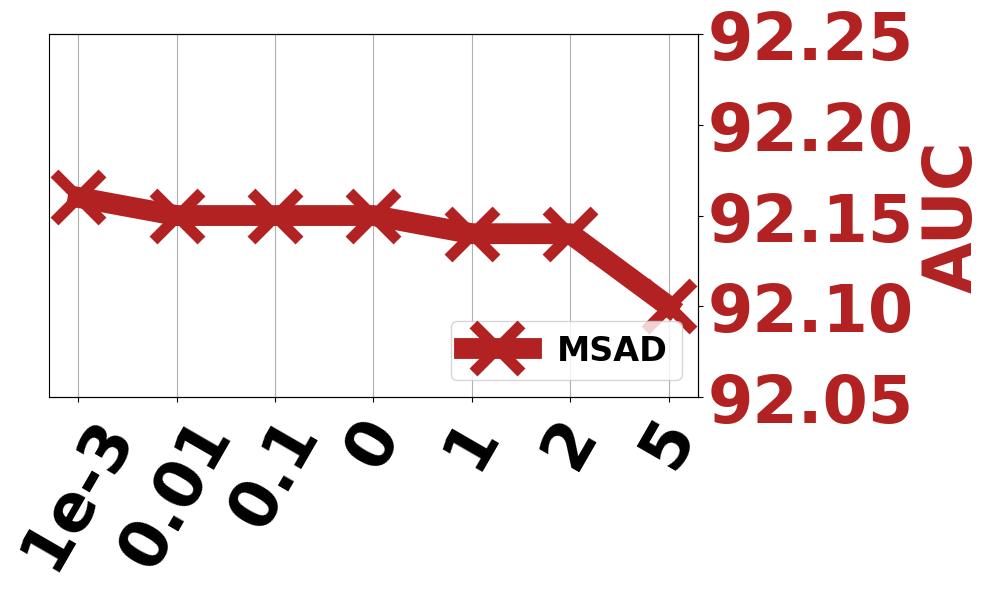}
\caption{$\beta$}
\label{hyper2}
\end{subfigure}
\begin{subfigure}[t]{0.242\linewidth}
\centering\includegraphics[trim=0cm 0cm 0cm 0cm, clip=true, width=\linewidth]{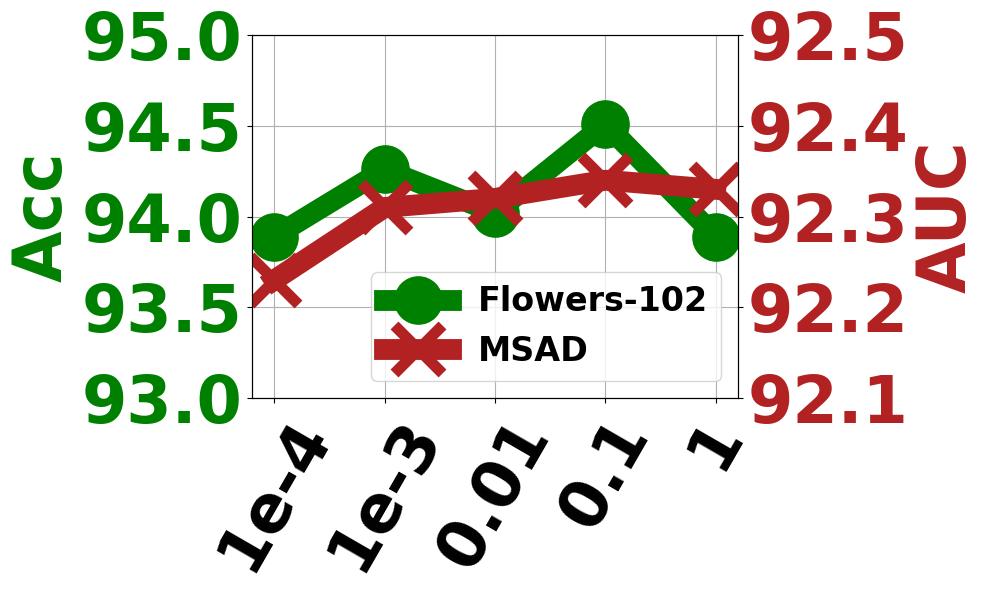}
\caption{$\gamma$}
\label{hyper3}
\end{subfigure}
\begin{subfigure}[t]{0.242\linewidth}
\centering\includegraphics[trim=0cm 0cm 0cm 0cm, clip=true, width=\linewidth]  {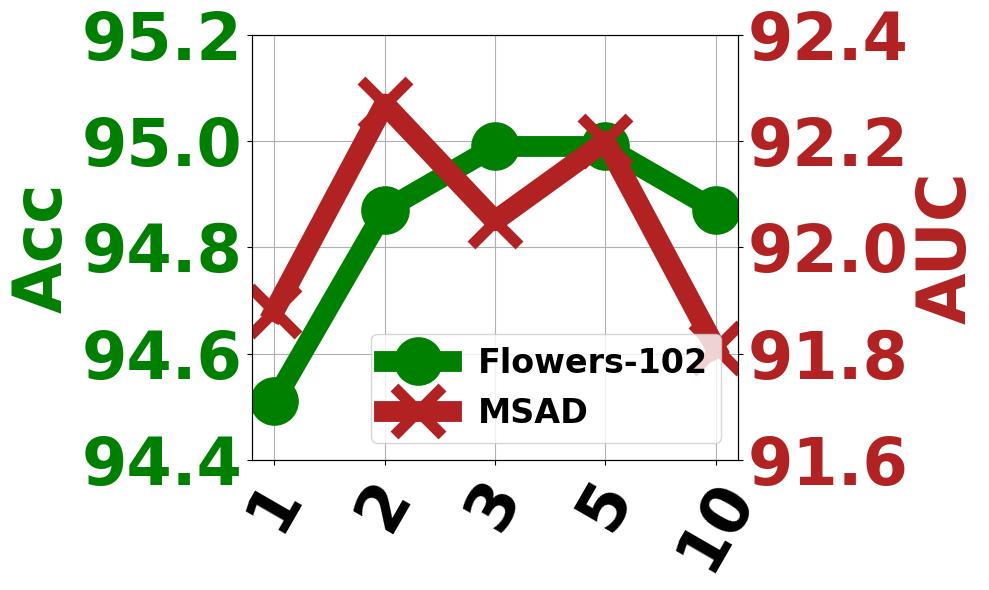} 
\caption{$N$}
\label{hyper4}
\end{subfigure}
\caption{Evaluation of key hyperparameters.}
\label{fig:hyper}
\end{figure}

\begin{figure}[tbp]
\centering
\begin{subfigure}[t]{0.495\linewidth}
\centering\includegraphics[trim=0cm 0cm 0cm 0cm, clip=true, width=\linewidth]{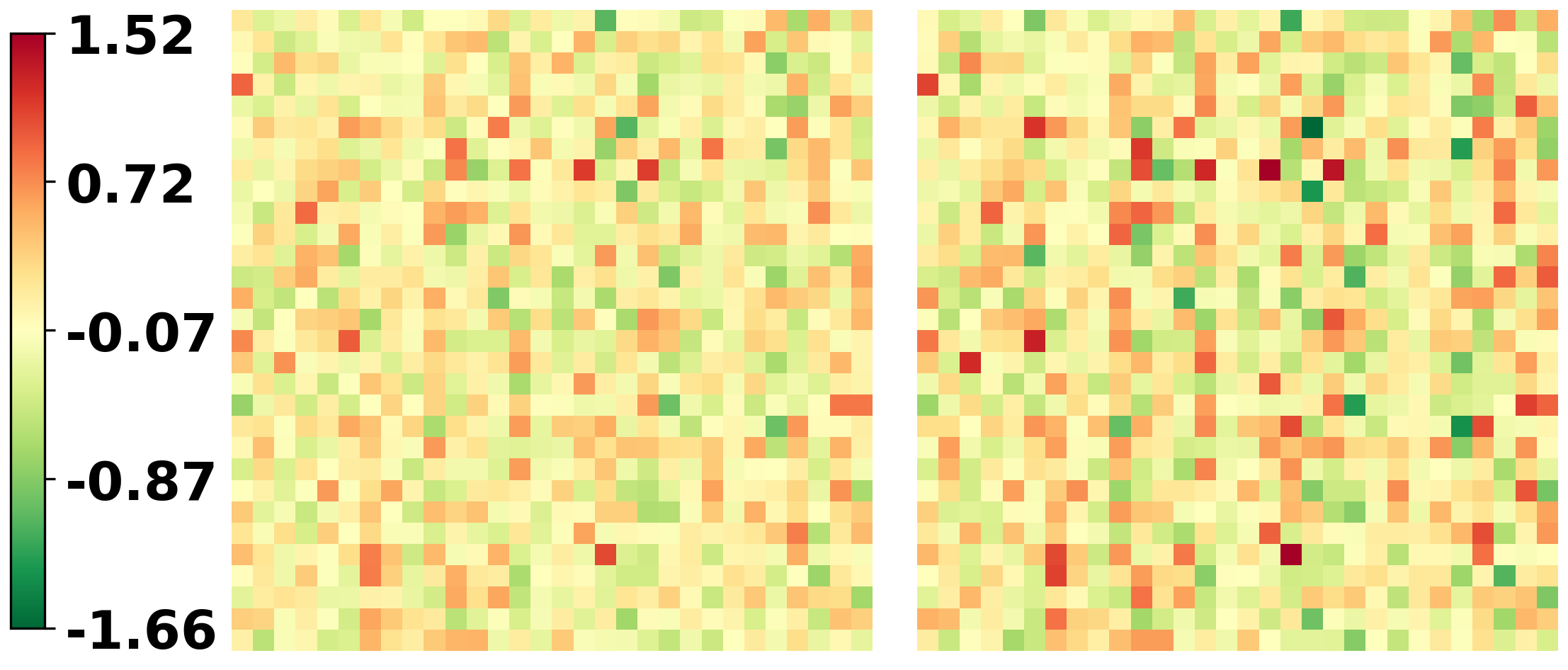}
\caption{Flowers-102}
\label{fig:flower-fc-vis}
\end{subfigure}
\begin{subfigure}[t]{0.495\linewidth}
\centering\includegraphics[trim=0cm 0cm 0cm 0cm, clip=true, width=\linewidth]{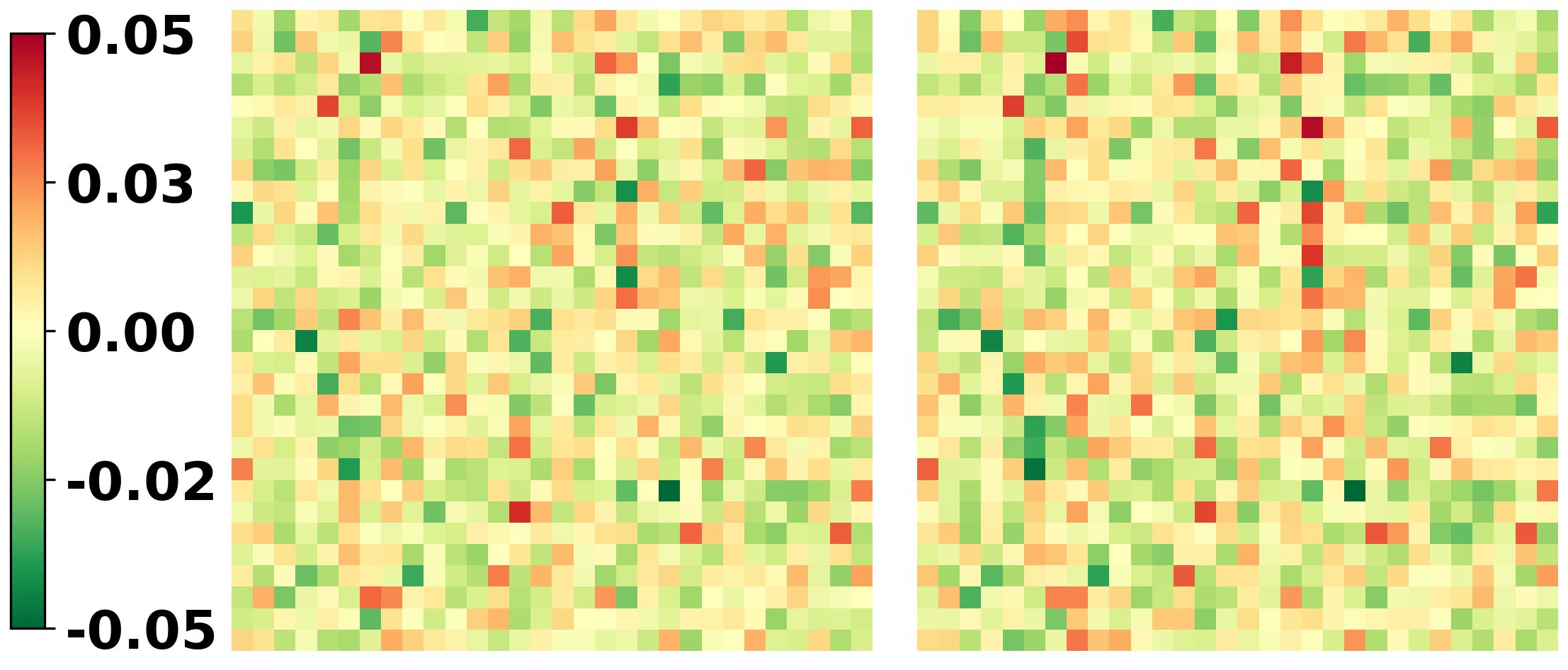}
\caption{SoyAging}
\label{fig:soy-fc-vis}
\end{subfigure}
\caption{
Visualization of selected FC weight regions shows a clear comparison between the baseline (left) and our temporal modeling (right). Temporal modeling yields stronger, more distinct patterns, enhancing feature discrimination. Even on the ultra-fine-grained SoyAging, our approach produces clearer, more structured weights, demonstrating the advantages of temporal supervision in feature learning.
}
\label{fig:fc-vis}
\end{figure}

\section{Experiment}

\subsection{Experimental Setup}

\textbf{Datasets.} We evaluate fine-grained recognition on Stanford Cars \cite{6755945}, Dogs \cite{KhoslaYaoJayadevaprakashFeiFei_FGVC2011}, Flowers-102 \cite{4756141}, and the ultra-fine-grained SoyAging dataset \cite{9710088}. Video anomaly detection uses MSAD \cite{zhu2024advancing} with Protocol ii, covering various anomaly types and scenarios. All evaluations follow standard protocols for fair comparison. 

\noindent\textbf{Models.} Our method trains a single FC on frozen vision transformer features, without fine-tuning the backbone. We extract features using CLIP-ViT-L/14/224 (ImageNet-1K) for Stanford Cars, CLIP-ViT-B/16/224 (ImageNet-1K) for Oxford Flowers-102, and ViT-B/16/224 (ImageNet-1K) for Stanford Dogs. For the specialized SoyAging dataset, we use CLE-ViT (Swin-B/448) pretrained on ImageNet-21K \cite{yu2023cle} to use its larger corpus for ultra-fine-grained tasks. For MSAD, frame-level features are extracted using CLIP-ViT-B/16/224 pretrained on WebImgText. 

\noindent\textbf{Setups.} In all experiments, the baseline is a static classifier with a single FC layer and softmax. We extend this by exploring feature trajectories, which represent smooth feature evolution through synthetic temporal augmentations (for static images) or natural temporal changes (for videos like MSAD). Our full model integrates feature trajectories with SEQ to effectively capture dynamic visual patterns over time.
%
We benchmark against recent state-of-the-art methods on each dataset to validate the competitiveness of our results.


\begin{table*}[tbp]
    \setlength{\tabcolsep}{0.15em}
    \renewcommand{\arraystretch}{0.70}
    \centering
    
    \resizebox{\linewidth}{!}{
    \begin{tabular}{ll
        cc cc cc cc
        cc cc cc cc
        cc cc cc cc
        }
    \toprule
    \multirow{3}{*}{Method} 
    & \multicolumn{2}{c}{Assault} & \multicolumn{2}{c}{Explosion} & \multicolumn{2}{c}{Fighting} & \multicolumn{2}{c}{Fire} 
    & \multicolumn{2}{c}{Obj. Fall} & \multicolumn{2}{c}{People Fall} & \multicolumn{2}{c}{Robbery} & \multicolumn{2}{c}{Shooting}
    & \multicolumn{2}{c}{Traffic Acc.} & \multicolumn{2}{c}{Vandalism} & \multicolumn{2}{c}{Water Inc.} & \multicolumn{2}{c}{\textbf{Overall}} \\
    \cmidrule(lr){2-3} \cmidrule(lr){4-5} \cmidrule(lr){6-7} \cmidrule(lr){8-9}
    \cmidrule(lr){10-11} \cmidrule(lr){12-13} \cmidrule(lr){14-15} \cmidrule(lr){16-17}
    \cmidrule(lr){18-19} \cmidrule(lr){20-21} \cmidrule(lr){22-23} \cmidrule(lr){24-25}
    & AUC & AP & AUC & AP & AUC & AP & AUC & AP 
    & AUC & AP & AUC & AP & AUC & AP & AUC & AP
    & AUC & AP & AUC & AP & AUC & AP & AUC & AP \\
    \midrule
    RTFM (I3D) 
    & 53.9 
    & \underline{\textbf{66.4}} 
    & 66.0 & 76.6 
    & 79.8 & 88.6 
    & 44.9 & 71.1 
    & 84.6 & 89.3 
    & 45.7 
    & \underline{\textbf{52.6}} 
    & \textbf{70.2} 
    & \underline{\textbf{88.0}} 
    & \underline{\textbf{87.5}} 
    & \underline{\textbf{89.2}} 
    & 64.1 
    & \textbf{57.7} 
    & 74.9 & 73.0 
    & 98.1 & 99.6 
    & 86.6 & 68.4 \\
    MGFN (SwinT) 
    & 50.2 & 49.6 
    & 50.9 & 58.1 
    & 57.2 & 67.1 
    & 51.4 & 74.2 
    & 41.3 & 51.6 
    & 44.4 & 40.3 
    & 40.1 & 68.5 
    & 51.4 & 63.9 
    & 50.4 & 42.3 
    & 42.6 & 40.9 
    & 58.6 & 87.2 
    & 69.3 & 33.6 \\
    MGFN (I3D) 
    & 53.9 & 60.2 
    & 59.1 & 66.5 
    & \textbf{80.6} & \textbf{89.5} 
    & 66.1 & 82.9 
    & 89.9 & 94.6 
    & \textbf{53.6} & 44.9 
    & \underline{\textbf{72.2}} 
    & 85.4 
    & 68.3 & 80.6 
    & \textbf{66.9} & 54.7 
    & 84.4 & 78.5 
    & 81.9 & 96.1 
    & 81.2 & 59.3 \\
    UR-DMU 
    & 56.9 
    & \textbf{64.5} 
    & 67.9 & 74.5 
    & \underline{\textbf{83.9}} 
    & \underline{\textbf{90.4}} 
    & 61.2 & 82.9 
    & \textbf{92.1} 
    & \underline{\textbf{95.8}} 
    & 42.5 & 43.7 
    & 63.5 & 79.3 
    & \textbf{81.4} & 87.8 
    & 62.0 & 55.6 
    & 84.7 & 77.0 
    & 98.5 & 99.5 
    & 85.0 & 68.3 \\
    EGO
    & 52.2 & 57.5 
    & 57.6 & 74.4 
    & 66.5 & 72.8 
    & 62.9 
    & \textbf{86.7} 
    & \underline{\textbf{92.3}} 
    & \textbf{94.8} 
    & 35.4 & 43.8 
    & 64.8 
    & \underline{\textbf{87.5}} 
    & 68.6 & 78.4 
    & 69.9 & \underline{\textbf{64.3}} 
    & \underline{\textbf{88.1}} 
    & \underline{\textbf{81.4}} 
    & 81.9 & 95.4 
    & 87.3 & 64.4 \\
    IEF-VAD 
    & \underline{\textbf{66.0}} & – 
    & 66.3 & – 
    & 79.8 & – 
    & 49.4 & – 
    & 75.9 & – 
    & 42.5 & – 
    & 66.9 & – 
    & \textbf{86.9} & – 
    & \underline{\textbf{70.1}} & – 
    & 75.8 & – 
    & \textbf{88.9} & – 
    & 82.1 & – \\

    \addlinespace[0.3ex]
    \hline
    \addlinespace[0.3ex]
    Baseline 
    & 48.2 & 51.5 
    & \textbf{77.3} & 84.6 
    & 73.1 & 83.9 
    & 78.2 & \textbf{94.7} 
    & 83.7 & 89.7 
    & 49.8 & 46.3 
    & 65.5 & 86.4 
    & 79.8 & 88.1 
    & 63.0 & 55.2 
    & 76.2 & 75.0 
    & \textbf{99.6} & \textbf{99.9} 
    & 86.7 & 72.2 \\
    \highlightrow
    \textit{w/ feat. traj.} 
    & 50.6 & 51.1 
    & 76.1 & \textbf{85.3} 
    & 70.5 & 82.8 
    & 76.7 & 94.5 
    & \textbf{85.6} & 90.7 
    & \underline{\textbf{56.2}} & \textbf{50.7} 
    & 67.0 & 86.2 
    & 79.1 & 88.3 
    & 61.4 & 52.6 
    & 84.3 & 78.3 
    & 99.3 & 99.8 
    & \underline{\textbf{92.1}} & \textbf{77.3} \\
    \highlightrow
    \textit{w/ feat. traj.\&SEQ} 
    & \textbf{59.3} & 60.2 
    & \underline{\textbf{84.9}} & \underline{\textbf{88.9}} 
    & 79.8 & \underline{\textbf{89.9}} 
    & \underline{\textbf{81.4}} & \underline{\textbf{95.6}} 
    & 85.2 & 90.8 
    & 52.3 & 48.7 
    & 68.1 & 87.4 
    & 79.5 & \textbf{88.6} 
    & 60.6 & 50.4 
    & \textbf{85.0} & \textbf{80.3} 
    & \underline{\textbf{99.6}} & \underline{\textbf{99.9}} 
    & \textbf{90.5} & \underline{\textbf{77.5}} \\

    \bottomrule
    \end{tabular}}
    
    \caption{Performance by anomaly type on MSAD.  
    The best result is marked with \underline{\textbf{bold and underline}}, and the second-best is shown in \textbf{bold}. We compare against recent methods, including RTFM \cite{Tian_2021_ICCV}, MGFN \cite{chen2023mgfn}, UR-DMU \cite{zhou2023dual}, EGO \cite{ding2025learnable}, and IEF-VAD \cite{jeong2025uncertainty}, which rely on complex architectures and spatio-temporal feature extraction. In contrast, our method, despite using only a temporal classifier on top of frozen features, consistently achieves strong and often superior results across anomaly types. This highlights the effectiveness of simple temporal modeling in capturing temporal dynamics without heavy architectural complexity.}
    \label{tab:anomaly-msad}
\end{table*}

\begin{table*}[tbp]
\setlength{\tabcolsep}{0.08em}
    \renewcommand{\arraystretch}{0.70}
    \centering
    
    \resizebox{\linewidth}{!}{
    \begin{tabular}{
        lcccccccccccccccccccccccccc
    }
    \toprule
    \multirow{3}{*}{Method}
    & \multicolumn{2}{c}{Frontdoor} &  \multicolumn{2}{c}{Mall} & \multicolumn{2}{c}{Office}
    & \multicolumn{2}{c}{Parkinglot} & \multicolumn{2}{c}{Pedestr. st.} & \multicolumn{2}{c}{Restaurant}
    & \multicolumn{2}{c}{Road} & \multicolumn{2}{c}{Shop} & \multicolumn{2}{c}{Sidewalk} & \multicolumn{2}{c}{St. highview}
    & \multicolumn{2}{c}{Train} & \multicolumn{2}{c}{Warehouse} & \multicolumn{2}{c}{\textbf{Overall}} \\
    \cmidrule(lr){2-3} \cmidrule(lr){4-5} \cmidrule(lr){6-7} \cmidrule(lr){8-9}
    \cmidrule(lr){10-11} \cmidrule(lr){12-13} \cmidrule(lr){14-15} \cmidrule(lr){16-17}
    \cmidrule(lr){18-19} \cmidrule(lr){20-21} \cmidrule(lr){22-23} \cmidrule(lr){24-25}
    \cmidrule(lr){26-27} 
    & AUC & AP & AUC & AP
    & AUC & AP & AUC & AP & AUC & AP & AUC & AP
    & AUC & AP & AUC & AP & AUC & AP & AUC & AP
    & AUC & AP & AUC & AP & AUC & AP \\
    \midrule

    RTFM (I3D) 
    & 81.8 & 79.3 
    & 88.1 & 76.6 
    & 76.6 & 72.8 
    & 80.7 & 45.8 
    & 94.0 & \textbf{48.5} 
    & 88.3 & 79.1 
    & \textbf{84.3} & 57.9 
    & 85.3 & 75.6 
    & \textbf{88.3} & 68.8 
    & 72.0 & 28.5 
    & 51.4 & 3.3 
    & 82.7 & 57.0 
    & 86.6 & 68.4 \\

    MGFN (SwinT) 
    & 59.5 & 51.7 
    & 18.5 & 20.1 
    & 64.1 & 52.3 
    & 67.9 & 19.0 
    & 75.9 & 9.7 
    & 67.9 & 44.0 
    & 70.6 & 26.3 
    & 62.7 & 43.0 
    & 69.0 & 25.9 
    & 75.3 & 23.3 
    & 65.4 & 5.2 
    & 70.1 & 30.1 
    & 69.3 & 33.6 \\

    MGFN (I3D) 
    & 82.5 & 80.8 
    & 73.8 & 71.3 
    & 71.5 & 58.2 
    & 68.9 & 14.8 
    & \textbf{94.8} & 36.2 
    & \underline{\textbf{95.1}} & \underline{\textbf{91.3}} 
    & 76.5 & 35.8 
    & 85.6 & 78.4 
    & 78.5 & 57.2 
    & 77.9 & 29.3 
    & 40.3 & 2.1 
    & 58.3 & 24.2 
    & 81.2 & 59.3 \\

    UR-DMU  
    & 84.8 & 82.8 
    & \underline{\textbf{91.0}} & \textbf{83.8} 
    & 77.8 & 67.3 
    & 91.4 & 53.9 
    & 81.9 & 11.5 
    & 93.1 & \textbf{87.4} 
    & 83.0 & \textbf{64.4} 
    & 81.3 & 64.5 
    & 86.5 & 64.1 
    & 85.0 & 37.7 
    & 59.0 & 3.1 
    & 81.2 & 59.1 
    & 85.0 & 68.3 \\

    EGO  
    & \textbf{85.2} & 81.6 
    & 82.3 & 73.4 
    & 80.0 & 71.7 
    & \textbf{96.8} & 75.2 
    & \underline{\textbf{97.5}} & \underline{\textbf{52.0}} 
    & \textbf{94.3} & 73.9 
    & \underline{\textbf{89.8}} & \underline{\textbf{64.6}} 
    & 83.4 & 72.2 
    & 87.1 & 45.0 
    & 28.2 & 10.1 
    & 80.8 & 7.8 
    & 84.7 & 46.6 
    & 87.3 & 64.4 \\

    IEF-VAD 
    & - & - & - & - & - & - & - & - & - & - & - & - & - & - & - & - & - & - & - & - & - & - & - & - & 82.1 & - \\

    \addlinespace[0.3ex]
    \hline
    \addlinespace[0.3ex]

    Baseline 
    & 83.9 & \textbf{83.3} 
    & \textbf{90.1} & 82.0 
    & 79.5 & \textbf{76.0} 
    & 96.3 & \textbf{83.9} 
    & 49.9 & 25.4 
    & 85.9 & 77.3 
    & 63.1 & 43.5 
    & \underline{\textbf{92.1}} & \underline{\textbf{84.7}} 
    & 85.4 & 67.8 
    & 99.7 & 98.8 
    & 91.7 & 24.2 
    & 79.0 & 45.9 
    & 86.7 & 72.2 \\
    \highlightrow
    \textit{w/ feat. traj.} 
    & \textbf{85.2} & 82.4 
    & \textbf{90.1} & 82.0 
    & \textbf{83.0} & 75.0 
    & \underline{\textbf{97.0}} & \underline{\textbf{86.3}} 
    & 46.8 & 10.7 
    & 90.8 & 81.5 
    & 81.4 & 60.4 
    & \textbf{90.9} & \textbf{82.0} 
    & \underline{\textbf{93.2}} & \underline{\textbf{80.9}} 
    & \underline{\textbf{99.9}} & \underline{\textbf{99.4}} 
    & \underline{\textbf{95.7}} & \underline{\textbf{42.9}} 
    & \underline{\textbf{94.2}} & \textbf{69.9} 
    & \underline{\textbf{92.1}} & \textbf{77.3} \\
    \highlightrow
    \textit{w/ feat. traj.\&SEQ}
    & \underline{\textbf{86.0}} & \underline{\textbf{84.6}} 
    & 89.6 & \underline{\textbf{84.2}} 
    & \underline{\textbf{84.5}} & \underline{\textbf{79.2}} 
    & 92.9 & 64.4 
    & 39.1 & 10.0 
    & 92.2 & 84.0 
    & 79.5 & 62.4 
    & 88.0 & 79.6 
    & 87.3 & \textbf{71.4} 
    & \textbf{99.8} & \textbf{99.2} 
    & \textbf{95.3} & \textbf{36.7} 
    & \textbf{93.5} & \underline{\textbf{70.9}} 
    & \textbf{90.5} & \underline{\textbf{77.5}} \\

    \bottomrule
    \end{tabular}
    }
    
    \caption{Performance by Scenario on MSAD. We report results on 12 test scenarios, excluding \textit{Highway} and \textit{Park}, which do not contain anomalous events.  
    Our methods, \textit{w/ feat. traj.} and \textit{w/ feat. traj.\&SEQ}, achieve strong performance across all scenarios. 
    }
    \label{tab:scenario-msad}
\end{table*}

\begin{figure}[tbp]
\centering
\begin{subfigure}[t]{0.495\linewidth}
\centering\includegraphics[trim=0cm 0cm 0cm 0cm, clip=true, width=\linewidth]{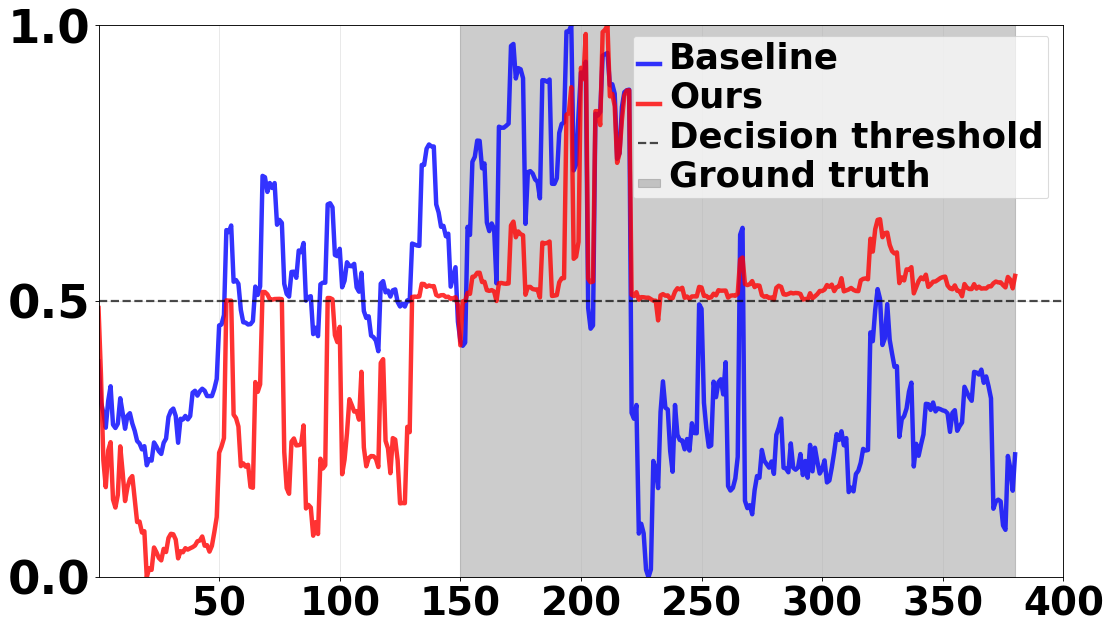}
\caption{Explosion}
\label{ano1}
\end{subfigure}
\begin{subfigure}[t]{0.495\linewidth}
\centering\includegraphics[trim=0cm 0cm 0cm 0cm, clip=true, width=\linewidth]{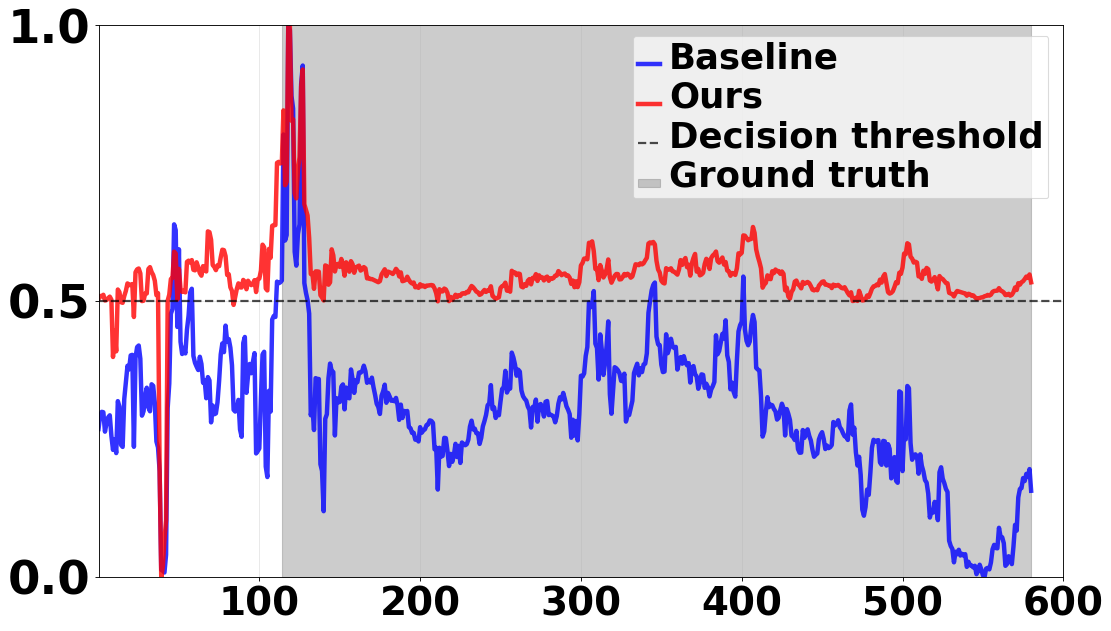}
\caption{People falling}
\label{ano2}
\end{subfigure}
\begin{subfigure}[t]{0.495\linewidth}
\centering\includegraphics[trim=0cm 0cm 0cm 0cm, clip=true, width=\linewidth]{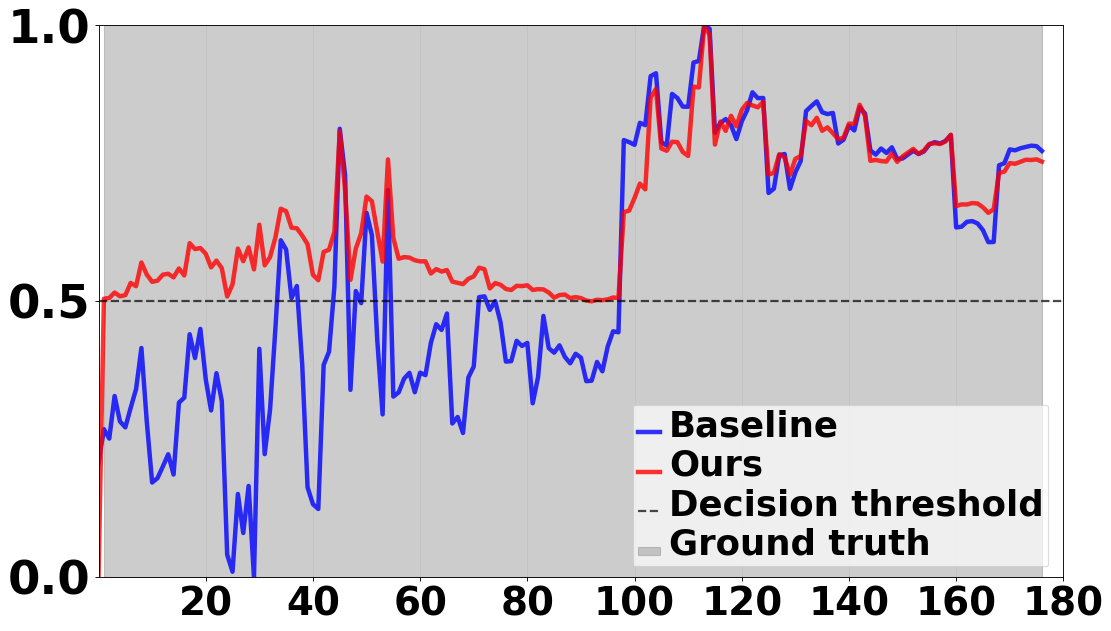}
\caption{Shooting}
\label{ano3}
\end{subfigure}
\begin{subfigure}[t]{0.495\linewidth}
\centering\includegraphics[trim=0cm 0cm 0cm 0cm, clip=true, width=\linewidth]{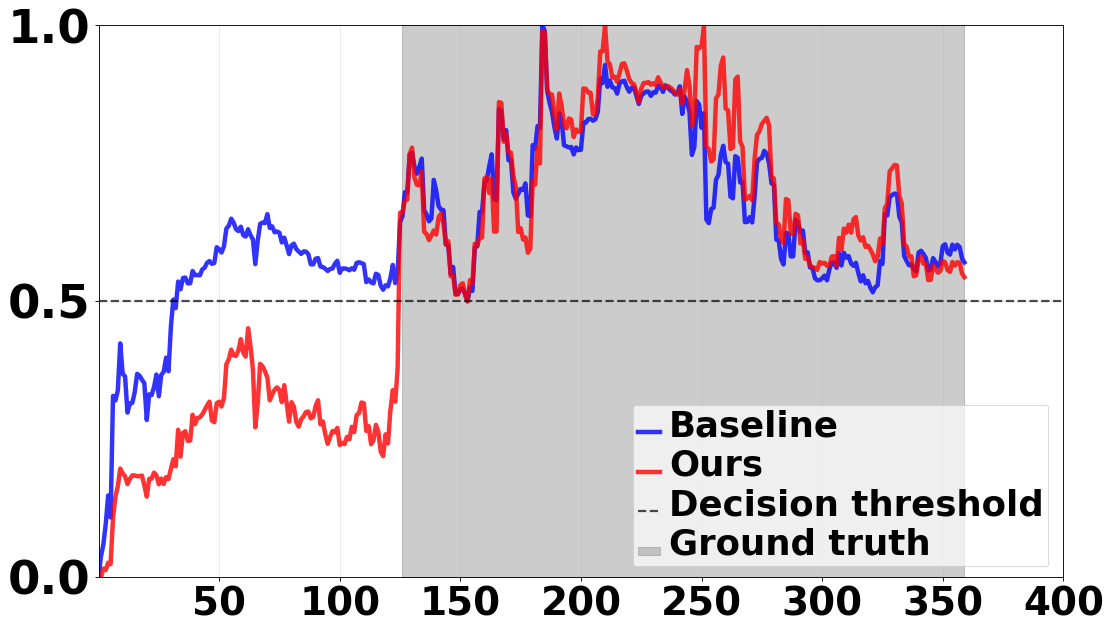} 
\caption{Traffic accident}
\label{ano4}
\end{subfigure}
\caption{Anomaly prediction comparison. Grey regions indicate ground-truth anomalies. Blue and red curves show the baseline and our method. Our approach detects anomalies more accurately and earlier, with scores crossing the 0.5 threshold in closer alignment with the ground truth.}
\label{fig:ano-vis}
\end{figure}

\subsection{Quantitative and Qualitative Evaluation}

\noindent\textbf{Hyperparameter evaluation.} 
Fig.~\ref{fig:hyper} shows the impact of key hyperparameters. The weight $\alpha$ controls the classification loss, with performance improving as $\alpha$ increases and stabilizing beyond $\alpha \geq 1$, underscoring the need to balance alignment and classification. The smoothness regularizer $\beta$ exhibits stable performance across a wide range, indicating robustness on MSAD. The softness parameter $\gamma$ in soft-DTW benefits from moderate values, \eg, $0.1$, particularly on Flowers-102. Lastly, $N$ defines the number of sequences in the support set for computing class exemplars. Larger $N$ leads to more reliable temporal estimates, with gains saturating at $N=3$ on Flowers-102.

\noindent\textbf{Analysis of learned weights.} 
As shown in Fig. \ref{fig:fc-vis}, temporal modeling produces stronger, more structured, and more distinct weight patterns compared to the baseline. These clearer weight structures suggest enhanced feature discrimination, as the model learns to better separate meaningful variations through temporal supervision.
Even on challenging ultra-fine-grained datasets such as SoyAging, where differences are inherently subtle, our method leads to more pronounced and organized weight patterns. This demonstrates that our approach introduces an implicit temporal inductive bias into otherwise static classifiers, without requiring architectural modifications. Instead, this bias is induced through simple temporal augmentations and sequence-based training objectives, providing a lightweight yet effective alternative to traditional heavy temporal models.
%

\noindent\textbf{Fine-grained image recognition.}
In Table \ref{tab:image}, when smooth temporal augmentations are applied, performance improves across all datasets. 
Adding SEQ learning on top of feature trajectories yields further improvements. This suggests that modeling temporal dependencies, even in static images, can enhance feature discrimination by implicitly learning consistent patterns and relationships across augmented views. Particularly in ultra-fine-grained recognition like SoyAging, the incremental gains indicate that temporal cues help address extreme subtlety in class differences where traditional spatial cues alone may be insufficient. 
These findings reveal that temporal consistency, typically associated with video or time-series data, can be used as a powerful inductive bias to improve static image classification, especially in challenging fine-grained domains, without increasing architectural complexity \cite{ding2025graph}. This insight opens new avenues for bridging temporal modeling techniques with static image tasks to push the limits of visual recognition accuracy.


\noindent\textbf{Evaluation on MSAD.} 
Analyzing performance by anomaly type (Table \ref{tab:anomaly-msad}), our method consistently achieves competitive or superior results compared to recent state-of-the-art approaches, despite using a simple classifier on frozen features. This demonstrates the effectiveness of lightweight temporal modeling in capturing complex anomaly dynamics without relying on heavy spatio-temporal architectures. Anomalies such as explosions, fires, and vandalism benefit notably from integrating feature trajectories and SEQ learning, leading to substantial performance gains (see Fig. \ref{fig:ano-vis} for prediction comparison).
The strong performance achieved without complex architectures further suggests that well-extracted frozen features, when paired with effective temporal modeling, provide a robust foundation for anomaly recognition.
Across scenarios (Table \ref{tab:scenario-msad}), the model maintains stable and reliable detection across diverse environments, from indoor settings like malls and offices to outdoor scenes such as sidewalks and parking lots. This demonstrates that temporal modeling via feature trajectories and SEQ learning adapts well to varied spatial contexts and activity patterns.



\noindent\textbf{On the role of \textit{time}.}
Our findings offer new insights into the importance of time in both visual and video recognition. Temporal information is not only crucial for video analysis but also enhances static image recognition when properly used. With simple temporal augmentations and SEQ-based supervision, standard feedforward classifiers, without any architectural modifications, can effectively reason over time. This demonstrates that temporal inductive bias can be introduced through training strategies rather than architectural complexity.
Our experiments consistently show performance gains across diverse datasets, from fine-grained image classification to video anomaly detection. These improvements stem from modeling feature trajectories and enforcing temporal consistency, enabling models to capture subtle dynamics and patterns over time. This lightweight approach \textit{challenges the common belief} that temporal reasoning requires heavy sequential models like transformers. Instead, our results emphasize that the way we supervise temporal information, by aligning it with feature evolution, can be equally or even more impactful.


\section{Conclusion}

We introduced a simple yet effective training framework that equips standard feedforward classifiers with temporal reasoning through a carefully designed loss function, without altering model architecture. At its core is our SEQ learning, which aligns prediction sequences with temporal prototypes via soft-DTW, further guided by semantic consistency and temporal smoothness objectives.
This framework enables lightweight classifiers to model temporal dynamics, yielding robust, temporally consistent predictions for both fine-grained visual recognition and video anomaly detection. By bridging static classifiers and temporal modeling through supervision alone, our method offers an efficient alternative to specialized sequence models. 


\section*{Acknowledgments}

Xi Ding, a visiting scholar at the ARC Research Hub for Driving Farming Productivity and Disease Prevention, Griffith University, conducted this work under the supervision of Lei Wang. 
Lei Wang proposed the algorithm and developed the theoretical framework, while Xi Ding implemented the code and performed the experiments.

We thank the anonymous reviewers for their invaluable insights and constructive feedback, which have contributed to improving our work.

This work was supported by the Australian Research Council (ARC) under Industrial Transformation Research Hub Grant IH180100002. 

This work was also supported by the National Computational Merit Allocation Scheme 2025 (NCMAS 2025; Lead CI: Lei Wang) and the ANU Merit Allocation Scheme (ANUMAS 2025; Lead CI: Lei Wang), with computational resources provided by NCI Australia, an NCRIS-enabled capability supported by the Australian Government.

\bibliography{aaai2026}

\newpage

\appendix

\maketitle

\section{Appendices}

\subsection{A. Video Anomaly Detection}

Video anomaly detection aims to identify abnormal events in temporal visual data. 
Conventional methods include reconstruction-based models \cite{7780455, ding2024quo, liu2018future, gong2019memorizing} and temporal embedding learning \cite{10.1145/3701716.3717746, park2020learning}, which often require specialized architectures and are sensitive to irrelevant changes like background motion or lighting \cite{wang2019loss,zhu2024advancing}.

Prototype-based anomaly detection has gained traction by modeling normality via latent prototypes \cite{huang2024prototype}. Our approach redefines normality in terms of prediction-space dynamics: deviations from class-consistent prediction trajectories indicate anomalies. This results in a lightweight, interpretable model capable of frame-level anomaly detection without complex reconstruction losses or custom temporal encoders.

\begin{tcolorbox}[width=1.0\linewidth, colframe=blackish, colback=beaublue, boxsep=0mm, arc=0mm, left=1mm, right=1mm, top=1mm, bottom=1mm]
We select image-pretrained backbones for several compelling reasons. First, they encode rich, generalizable visual representations transferable across diverse tasks. Second, unlike video-pretrained models that may overfit to motion-specific artifacts, ImageNet embeddings remain stable and domain-agnostic. Third, freezing the backbone yields substantial efficiency gains, reducing training time, memory footprint, and energy consumption, while mitigating overfitting risks. This modularity allows the classifier to concentrate exclusively on learning temporal relationships from high-quality visual features.
\end{tcolorbox}

\subsection{B. Temporal Inductive Bias in Action}
\begin{tcolorbox}[width=1.0\linewidth, colframe=blackish, colback=beaublue, boxsep=0mm, arc=0mm, left=1mm, right=1mm, top=1mm, bottom=1mm]
We present a principled and versatile strategy for embedding temporal inductive biases into standard feedforward classifiers. By encouraging models to learn not only class semantics but also the expected temporal evolution of predictions under smooth input variations, our method enables temporal reasoning without architectural modifications or reliance on sequence models. This approach bridges both static and temporal domains, from fine-grained visual recognition enhanced with synthetic dynamics to naturally evolving video streams, all while maintaining simplicity and broad applicability.
\end{tcolorbox}
\textbf{Fine-grained image recognition.}
In static recognition tasks, we simulate temporal progression by applying smoothly varying augmentations to individual images, such as gradual pose shifts, lighting changes, or appearance perturbations. These transformations create synthetic sequences that expose the classifier to the kinds of structured variations that occur in real-world observations. Through this process, the model learns to maintain consistent and confident predictions over these evolving inputs, effectively acquiring robustness to structured perturbations and intra-class variability, key challenges in fine-grained recognition. Importantly, our method captures not only static class identity but also characteristic prediction trajectories over these pseudo-temporal sequences, serving as an implicit form of temporal supervision even in datasets lacking real temporal signals.

\textbf{Video anomaly detection.} 
For temporal anomaly detection, we use naturally evolving video sequences to model the typical temporal dynamics of normal behavior. Support sets comprising normal sequences are used to construct prototype trajectories, which capture class-consistent temporal evolution in feature space. The model then performs fine-grained, frame-level anomaly detection by identifying deviations from these learned prototypes. This setup naturally supports early anomaly detection: anomalies can be flagged promptly as soon as prediction trajectories deviate from the expected normal patterns. Such responsiveness is critical in applications requiring timely monitoring and intervention, such as industrial inspection, medical monitoring, or security surveillance.

\subsection{C. Relation to Existing Frameworks}

Our SEQ learning adapts and extends concepts from contrastive learning, few-shot learning, and metric learning to provide a new framework for fine-grained visual recognition and detecting anomalous patterns in sequential data.

\textbf{Contrastive learning framework.} Our SEQ learning framework draws inspiration from contrastive learning, which learns robust representations by encouraging similarity between positive pairs and dissimilarity between negatives \cite{chen2020simple,kuang2021video,oord2018representation}. However, unlike conventional instance-level contrastive learning, SEQ operates at the sequence level, using temporal alignment rather than pointwise distance. It introduces an implicit temporal inductive bias by aligning prediction trajectories, promoting consistency across samples that share underlying class semantics and temporal evolution.
\begin{tcolorbox}[width=1.0\linewidth, colframe=blackish, colback=beaublue, boxsep=0mm, arc=0mm, left=1mm, right=1mm, right=1mm, top=1mm, bottom=1mm]
\begin{remark}
In SEQ learning, positive pairs are sequences from the same class, such as normal videos in anomaly detection or temporally-augmented samples from the same fine-grained category. The model minimizes alignment cost between these sequences, learning class-consistent temporal dynamics without requiring negative sampling.
\end{remark}
\end{tcolorbox}

Crucially, SEQ avoids explicit negative pairs and instead focuses on capturing intra-class temporal consistency. In fine-grained recognition, it models the subtle trajectory shifts introduced by smooth temporal augmentations, enabling static classifiers to learn temporal structure. In video anomaly detection, it captures the diverse but coherent normal behaviors, improving robustness and enabling early anomaly detection. This class-conditional sequence-level supervision offers a lightweight yet powerful alternative to conventional contrastive or recurrent models, especially in domains where temporal labels are scarce.

\textbf{Few-shot learning framework.} Our training strategy adopts an episodic structure reminiscent of few-shot learning, where each episode includes a query sequence and a support set drawn from the same class. This design encourages the model to generalize from limited supervision by learning class-specific patterns from few examples.
\begin{tcolorbox}[width=1.0\linewidth, colframe=blackish, colback=beaublue, boxsep=0mm, arc=0mm, left=1mm, right=1mm, right=1mm, top=1mm, bottom=1mm]
\begin{remark}
Unlike standard few-shot learning, SEQ learning dynamically samples diverse support instances per episode, forming class prototypes that capture intra-class variation. In fine-grained image recognition, support sets are sampled from temporally augmented views of the same class, fostering rich, class-conditional dynamics. For video anomaly detection, the model aligns anomalous query sequences with normal support sets to detect deviations, using the exemplar as a learned prototype of normal behavior.
\end{remark}
\end{tcolorbox}

This episodic formulation allows the model to build robust temporal prototypes from few but diverse samples, crucial in scenarios like anomaly detection, where abnormal events are rare and hard to enumerate \cite{zhu2024advancing}. By training to generalize from limited examples and learning deviations from normality, our approach naturally fits the few-shot paradigm while using temporal alignment for stronger inductive bias.

\textbf{Metric learning framework.} At the heart of our approach lies a differentiable metric, $\gamma$-soft-DTW \cite{cuturi2017soft}, which enables alignment-based comparison of sequences with varying lengths.
\begin{tcolorbox}[width=1.0\linewidth, colframe=blackish, colback=beaublue, boxsep=0mm, arc=0mm, left=1mm, right=1mm, right=1mm, top=1mm, bottom=1mm]
\begin{remark}
Our SEQ framework uses metric learning to minimize alignment cost between positive sequence pairs. In fine-grained recognition, queries are pulled closer to their class prototypes. In anomaly detection, sequences with normal behavior are aligned to form consistent, low-distance pairs.
\end{remark}
\end{tcolorbox}

Unlike traditional DTW, $\gamma$-soft-DTW is fully differentiable, enabling end-to-end optimization. It not only guides training via smooth gradients but also yields interpretable alignment matrices, revealing well-matched (or misaligned) temporal segments, useful for both classification and anomaly localization.

By embedding sequence alignment within a metric learning framework, SEQ effectively captures temporal similarity, offering a general and lightweight solution for tasks with limited labeled data.

\subsection{D. Proof of Smoothness Preservation}
\label{sec:smoothness_proof}

We formally prove that temporal smoothness in the feature space is preserved when the features are passed through a single fully connected (FC) layer followed by a Softmax or Sigmoid activation. This ensures that temporally adjacent features yield prediction scores that do not fluctuate abruptly, which is crucial for tasks such as video anomaly detection and temporally consistent fine-grained recognition.

Let $\mZ = [\vz_1, \vz_2, \dots, \vz_\tau] \in \mathbb{R}^{\tau \times d}$ denote a sequence of $d$-dimensional feature vectors ordered in time, where $\tau$ is the temporal length of the sequence. We assume that the feature sequence is temporally smooth in the $\ell_2$ norm:
\begin{equation}
    \|\vz_{i+1} - \vz_i\|_2 \leq \epsilon, \quad \forall i = 1, \dots, \tau \!-\! 1,
\end{equation}
for a small constant $\epsilon > 0$.

Let the classifier be defined as:
\begin{equation}
    \vphi_i = \sigma(\mW \vz_i + \vb),
\end{equation}
where $\mW \in \mathbb{R}^{C \times d}$ is the weight matrix, $\vb \in \mathbb{R}^C$ is the bias vector, $\vphi_i \in \mathbb{R}^C$ is the output vector of class scores (probabilities), and $\sigma: \mathbb{R}^C \to \mathbb{R}^C$ is either the Softmax or Sigmoid activation.

We aim to show that the prediction score sequence $\mPhi = [\vphi_1, \dots, \vphi_\tau]$ inherits the temporal smoothness of the feature sequence $\mZ$. That is, we want to prove that:
\begin{equation}
    \|\vphi_{i+1} - \vphi_i\|_2 \leq \delta, \quad \forall i = 1, \dots, \tau \!-\! 1,
\end{equation}
for some small $\delta > 0$ that depends on $\epsilon$ and the classifier parameters.

To establish this, we proceed with the following steps.

\textbf{Step 1: linearity of the fully connected layer.}

The linear transformation $g(\vz) = \mW \vz + \vb$ is Lipschitz continuous with Lipschitz constant $\|\mW\|_2$, where $\|\mW\|_2$ denotes the spectral norm of $\mW$:
\begin{equation}
    \|g(\vz_{i+1}) \!-\! g(\vz_i)\|_2 \!=\! \|\mW (\vz_{i+1} \!-\! \vz_i)\|_2 \!\leq\! \|\mW\|_2 \cdot \|\vz_{i+1} \!-\! \vz_i\|_2.
\end{equation}
Since $\|\vz_{i+1} - \vz_i\|_2 \leq \epsilon$, it follows that:
\begin{equation}
    \|g(\vz_{i+1}) - g(\vz_i)\|_2 \leq \|\mW\|_2 \cdot \epsilon.
\end{equation}

Let us define $\vu_i = \mW \vz_i + \vb \in \mathbb{R}^C$, so $\vphi_i = \sigma(\vu_i)$. We now analyze the effect of applying the activation $\sigma$ to the vector $\vu_i$.

\textbf{Step 2: smoothness of the activation function.}

Let us consider both Softmax and Sigmoid separately.

\textit{Case 1: Sigmoid.}  
The Sigmoid function applied element-wise is known to be $\frac{1}{4}$-Lipschitz. For any two inputs $\vu_i, \vu_{i+1} \in \mathbb{R}^C$:
\begin{equation}
    \|\sigma(\vu_{i+1}) - \sigma(\vu_i)\|_2 \leq \frac{1}{4} \cdot \|\vu_{i+1} - \vu_i\|_2.
\end{equation}
Combining with the result of Step 1:
\begin{equation}
    \|\vphi_{i+1} - \vphi_i\|_2 \leq \frac{1}{4} \cdot \|\mW\|_2 \cdot \epsilon.
\end{equation}

\textit{Case 2: Softmax.}  
The Softmax function is also Lipschitz continuous under $\ell_2$ norm. It is known that:
\begin{equation}
    \|\text{Softmax}(\vu_{i+1}) - \text{Softmax}(\vu_i)\|_2 \leq L_{\text{softmax}} \cdot \|\vu_{i+1} - \vu_i\|_2,
\end{equation}
where $L_{\text{softmax}} \leq 1$ and depends on the norm and input scale. Therefore:
\begin{equation}
    \|\vphi_{i+1} - \vphi_i\|_2 \leq L_{\text{softmax}} \cdot \|\mW\|_2 \cdot \epsilon.
\end{equation}

\textbf{Step 3: conclusion.}

In both cases, the final inequality takes the form:
\begin{equation}
    \|\vphi_{i+1} - \vphi_i\|_2 \leq L \cdot \epsilon,
\end{equation}
where $L = \frac{1}{4} \|\mW\|_2$ for Sigmoid activation, and $L = L_{\text{softmax}} \cdot \|\mW\|_2$ for Softmax activation.

This proves that the transition from temporally smooth features to prediction scores via a fully connected classifier followed by a smooth activation preserves temporal smoothness. That is, the classifier does not introduce abrupt changes in the predicted scores when the input features change gradually over time. This property is critical for tasks involving temporal modeling, where consistency across frames or clips must be maintained.

\begin{figure*}[tbp]
  \centering
  \begin{subfigure}[t]{0.8\textwidth}
    \centering
    \includegraphics[
      trim=0 0 0 0,clip,
      width=\linewidth,
      height=0.32\textheight,
      keepaspectratio
    ]{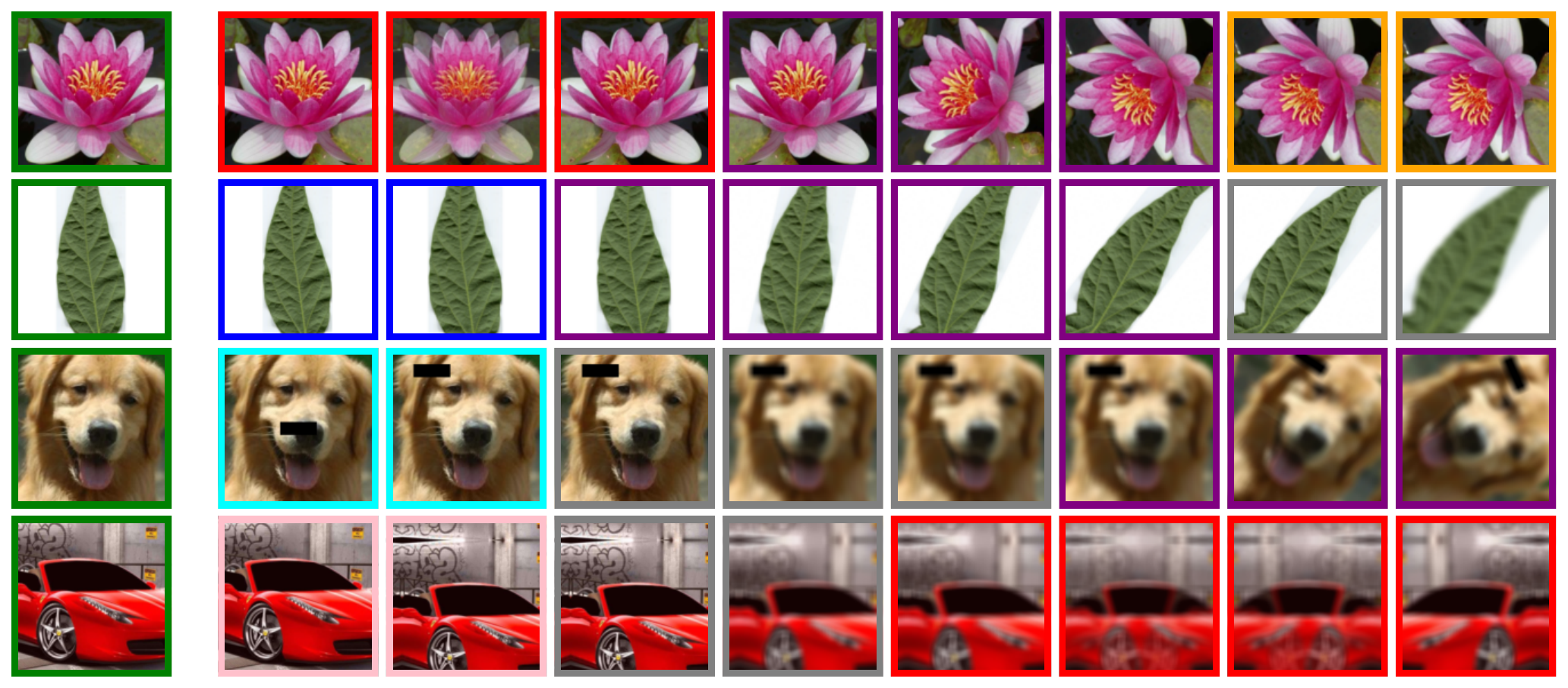}
    \label{fig:aug1}
  \end{subfigure}

  \vspace*{0.5em}  

  \begin{subfigure}[t]{0.8\textwidth}
    \centering
    \includegraphics[
      trim=0 0 0 0,clip,
      width=\linewidth,
      height=0.32\textheight,
      keepaspectratio
    ]{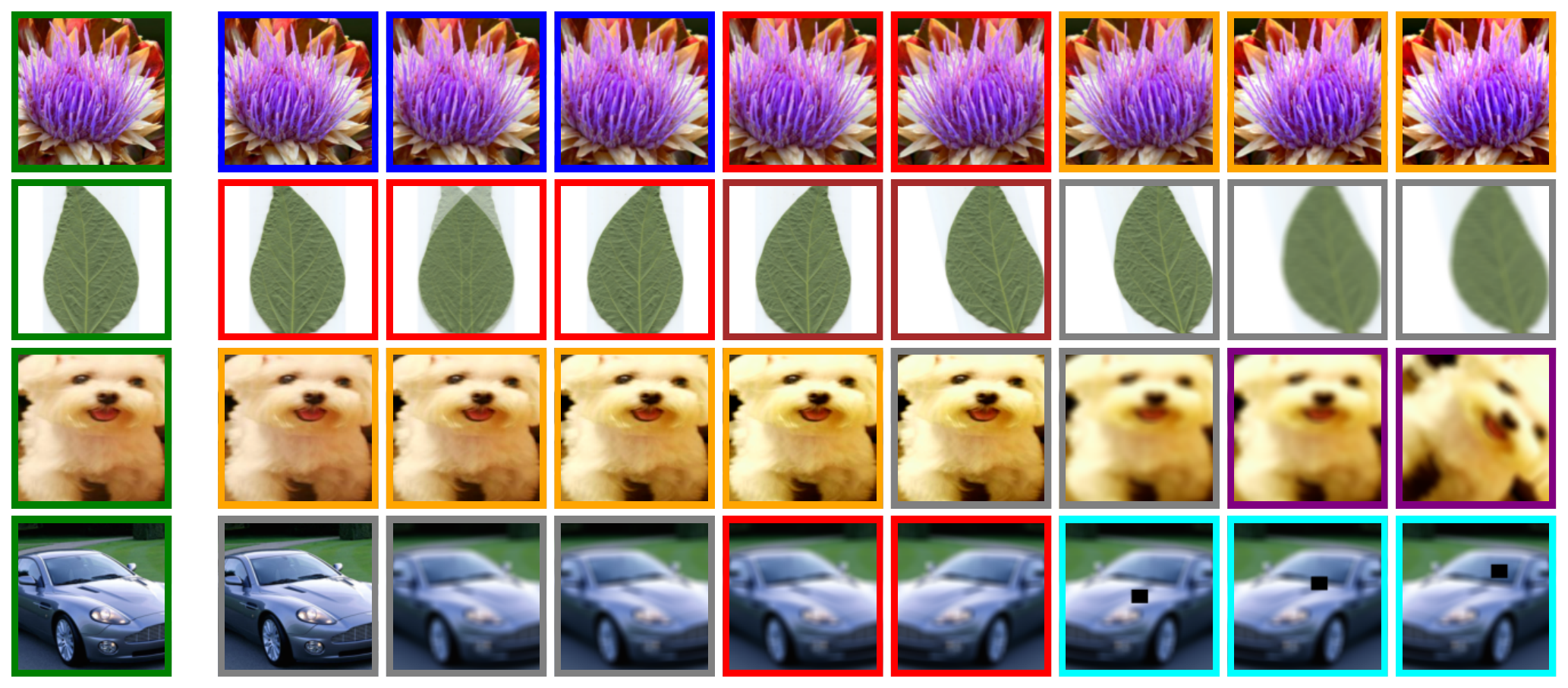}
    \label{fig:aug2}
  \end{subfigure}

  \vspace*{0.5em}

  \begin{subfigure}[t]{0.8\textwidth}
    \centering
    \includegraphics[
      trim=0 0 0 0,clip,
      width=\linewidth,
      height=0.32\textheight,
      keepaspectratio
    ]{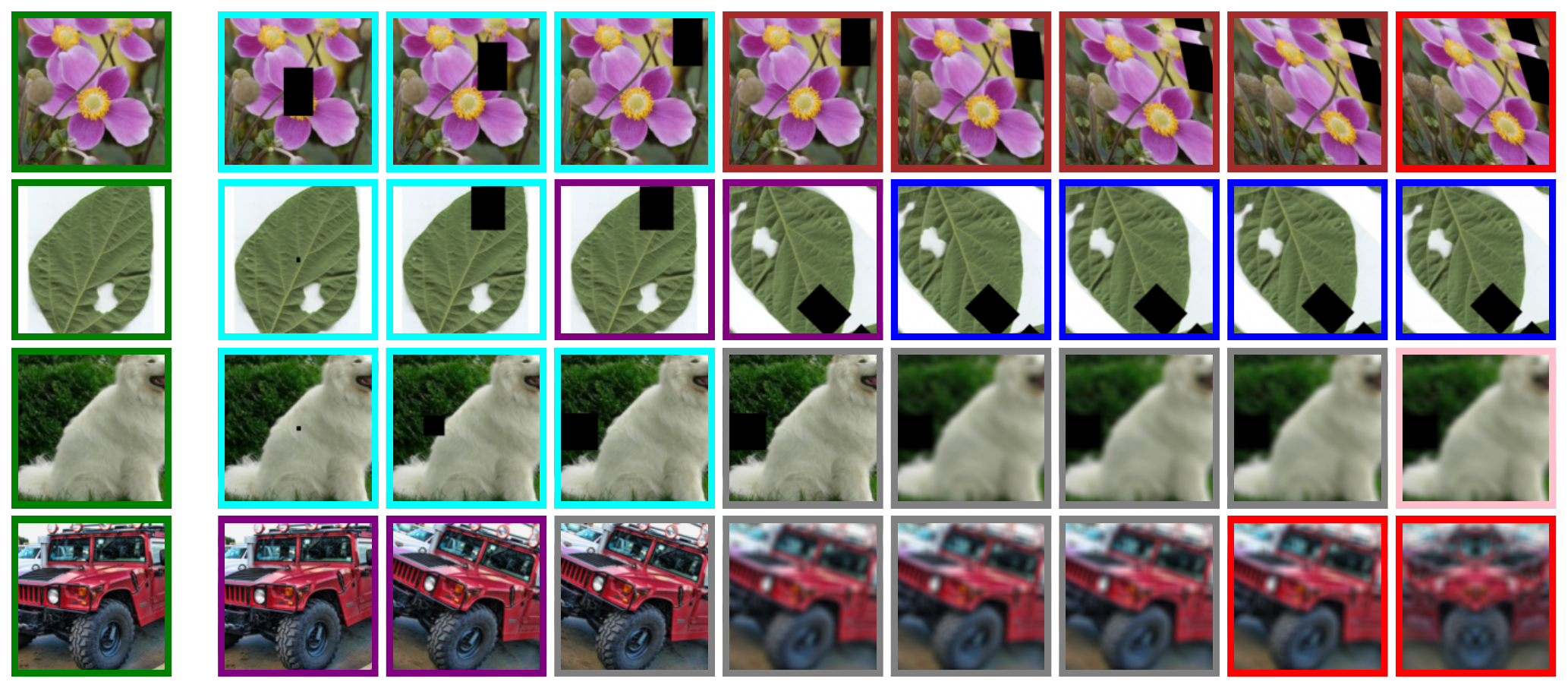}
    \label{fig:aug3}
  \end{subfigure}
  \caption{Examples from Flowers-102, SoyAging, Stanford Dogs, and Cars show how augmentations create temporal variations from one image. The first column shows originals (green); others apply augmentations by color: flip (red), zoom (blue), rotation (purple), color jitter (orange), shear (brown), translation (pink), blur (gray), and cutout (cyan), enriching the feature space with varied appearances.}
  \label{fig:vertical_time_aug1}
\end{figure*}

\begin{figure*}[tbp]
\ContinuedFloat
  \centering

  \begin{subfigure}[t]{0.8\textwidth}
    \centering
    \includegraphics[
      trim=0 0 0 0,clip,
      width=\linewidth,
      height=0.32\textheight,
      keepaspectratio
    ]{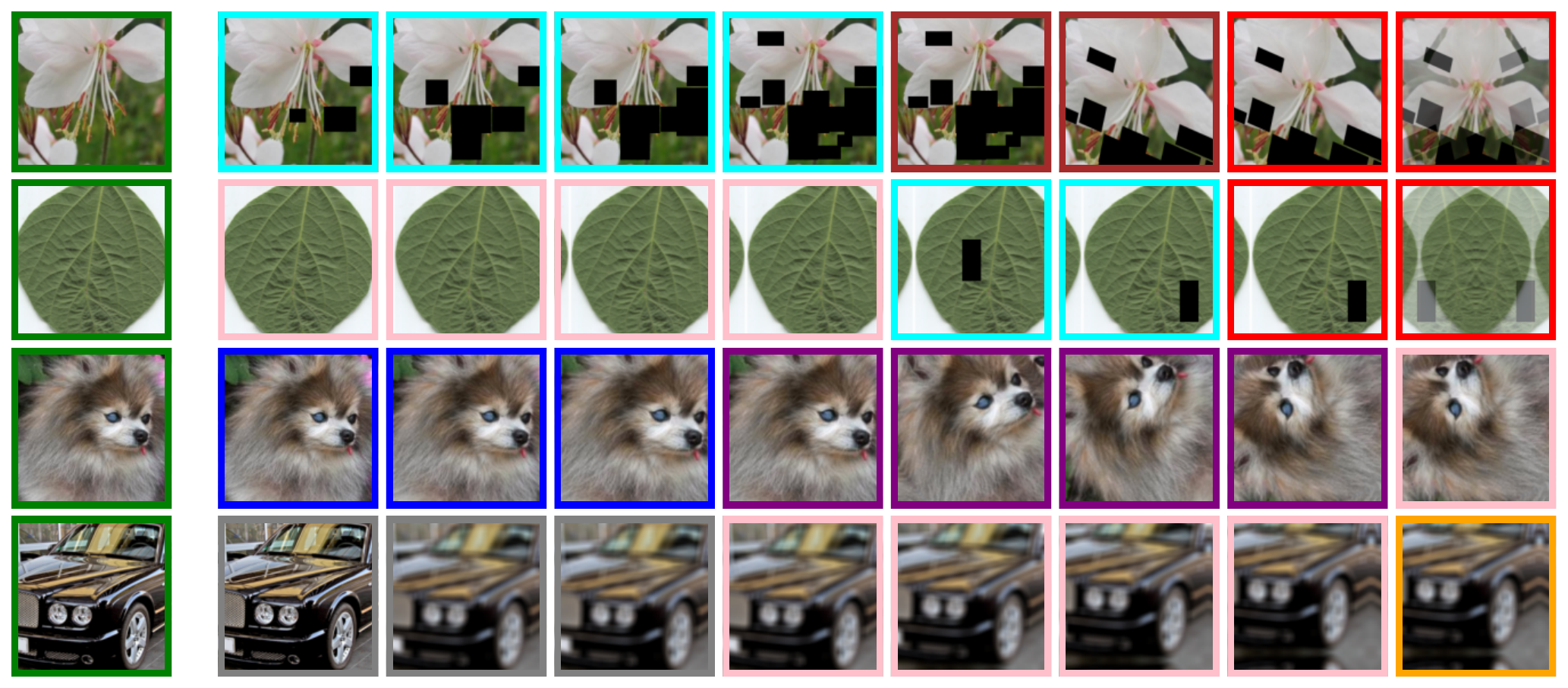}
    \label{fig:aug4}
  \end{subfigure}

  \vspace*{0.5em}  

  \begin{subfigure}[t]{0.8\textwidth}
    \centering
    \includegraphics[
      trim=0 0 0 0,clip,
      width=\linewidth,
      height=0.32\textheight,
      keepaspectratio
    ]{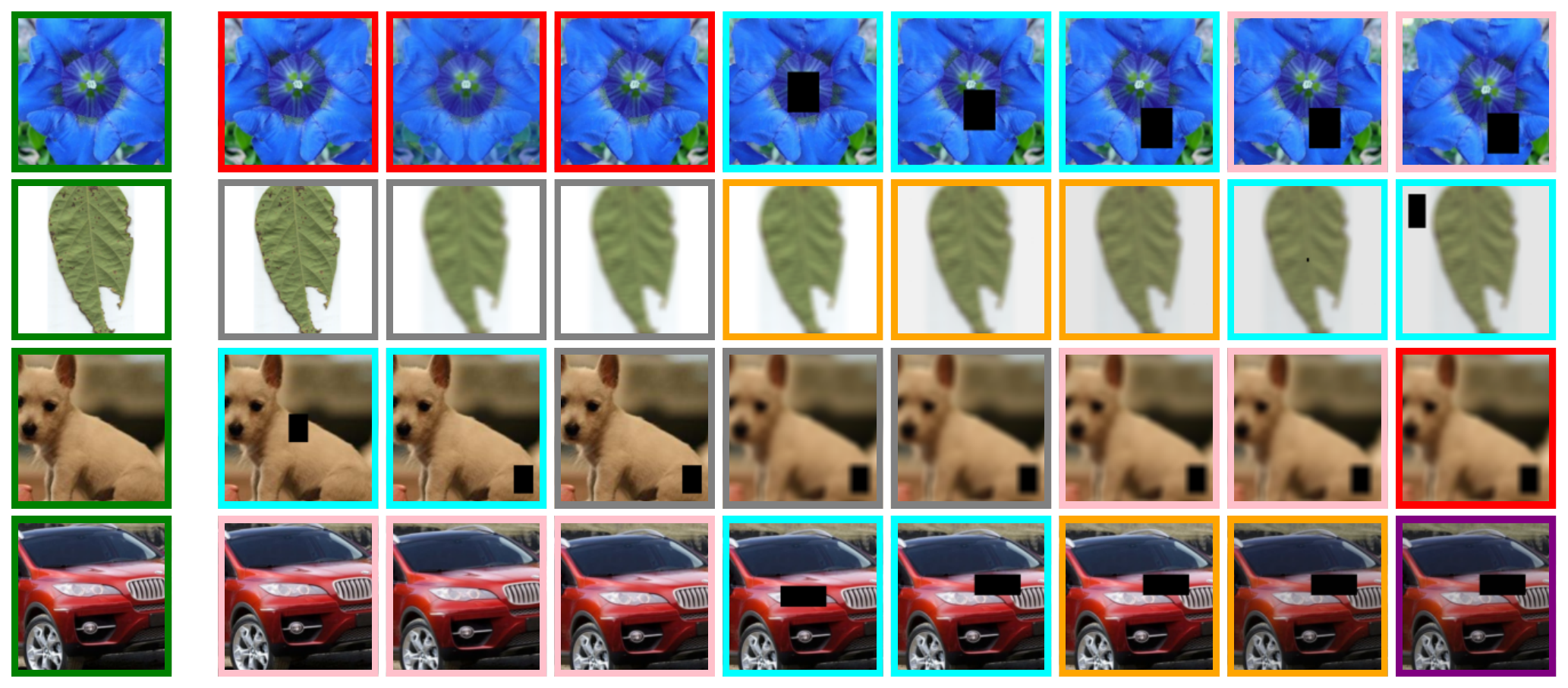}
    \label{fig:aug6}
  \end{subfigure}

  \vspace*{0.5em}

  \begin{subfigure}[t]{0.8\textwidth}
    \centering
    \includegraphics[
      trim=0 0 0 0,clip,
      width=\linewidth,
      height=0.32\textheight,
      keepaspectratio
    ]{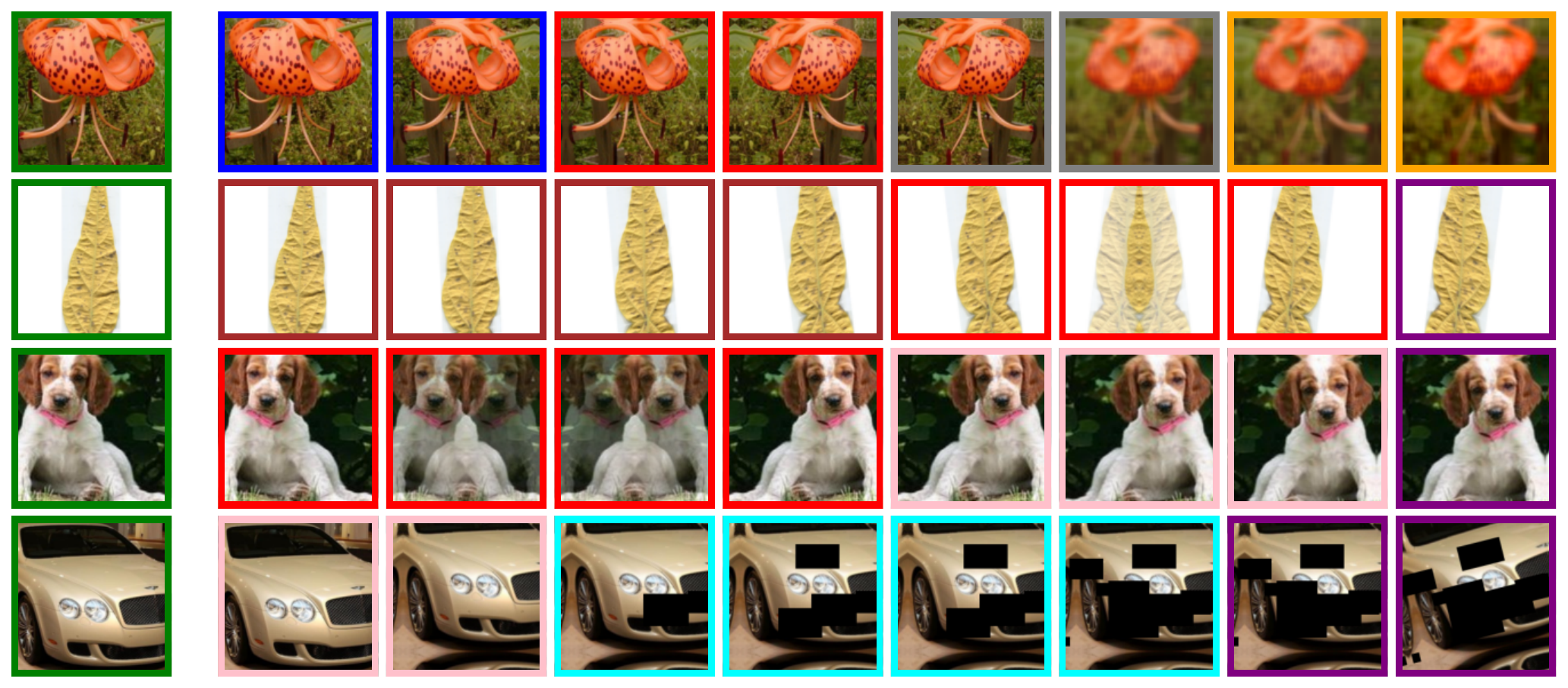}
    \label{fig:aug7}
  \end{subfigure}
  \caption{(continued) Examples from Flowers-102, SoyAging, Stanford Dogs, and Cars show how augmentations create temporal variations from one image. The first column shows originals (green); others apply augmentations by color: flip (red), zoom (blue), rotation (purple), color jitter (orange), shear (brown), translation (pink), blur (gray), and cutout (cyan), enriching the feature space with varied appearances.}
  \label{fig:vertical_time_aug2}
\end{figure*}

\begin{figure*}[tbp]
\ContinuedFloat
  \centering

  \begin{subfigure}[t]{0.8\textwidth}
    \centering
    \includegraphics[
      trim=0 0 0 0,clip,
      width=\linewidth,
      height=0.32\textheight,
      keepaspectratio
    ]{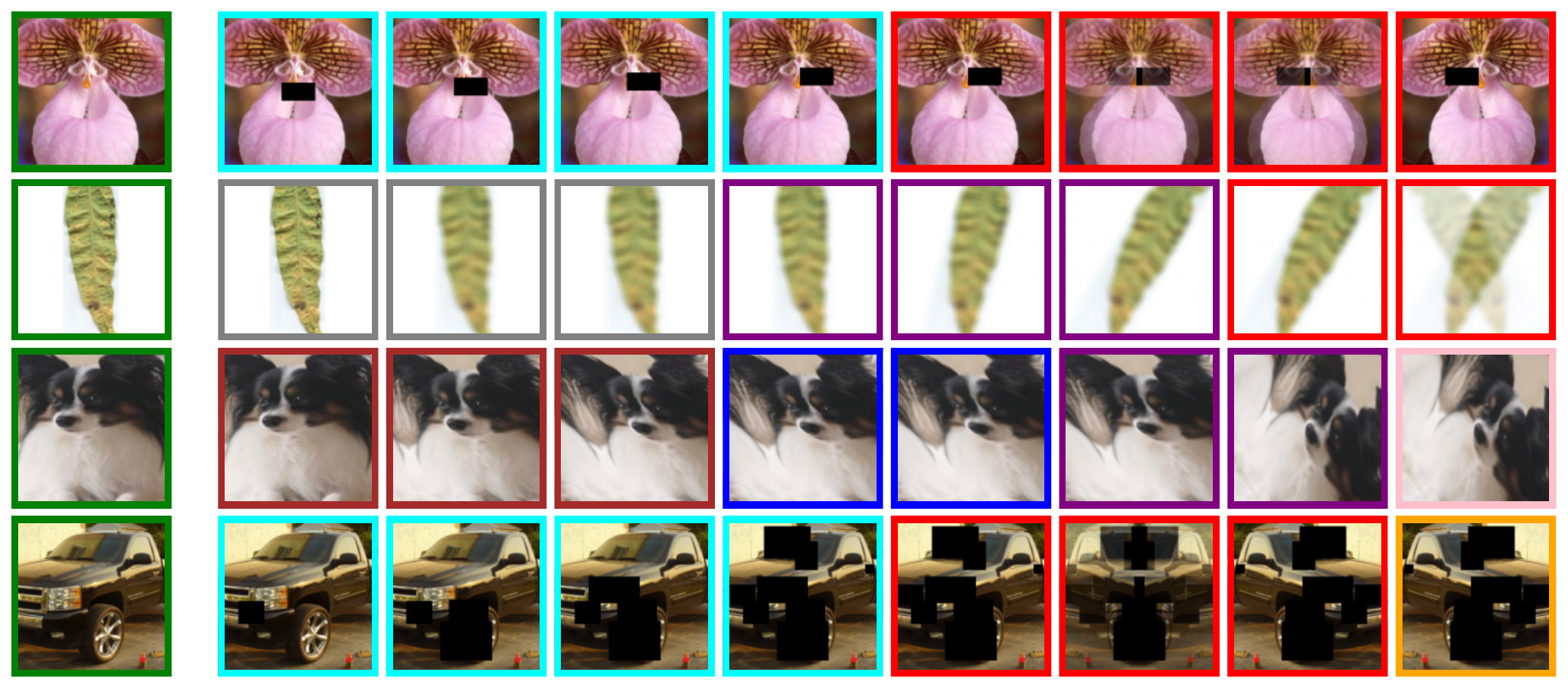}
    \label{fig:aug8}
  \end{subfigure}

  \vspace*{0.5em}  

  \begin{subfigure}[t]{0.8\textwidth}
    \centering
    \includegraphics[
      trim=0 0 0 0,clip,
      width=\linewidth,
      height=0.32\textheight,
      keepaspectratio
    ]{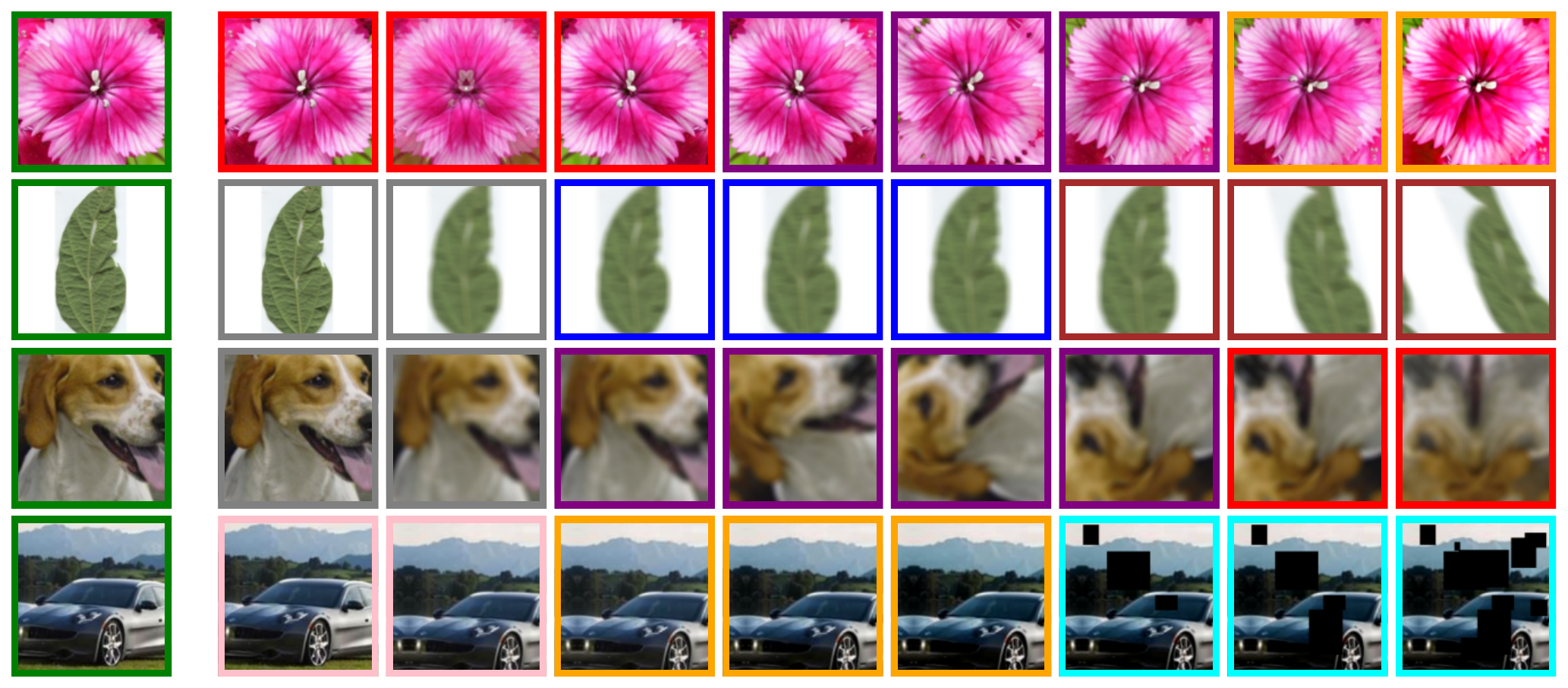}
    \label{fig:aug9}
  \end{subfigure}

  \vspace*{0.5em}

  \begin{subfigure}[t]{0.8\textwidth}
    \centering
    \includegraphics[
      trim=0 0 0 0,clip,
      width=\linewidth,
      height=0.32\textheight,
      keepaspectratio
    ]{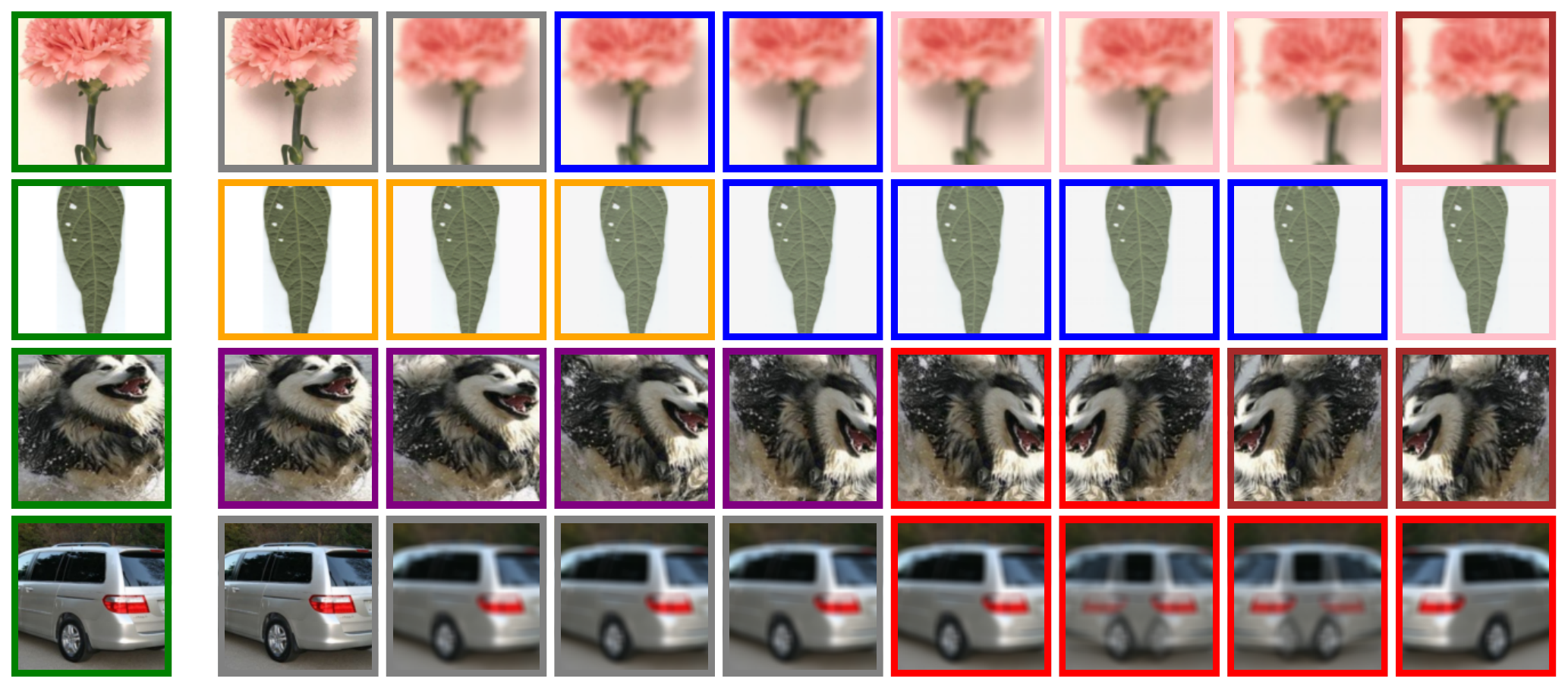}
    \label{fig:aug10}
  \end{subfigure}
  \caption{(continued) Examples from Flowers-102, SoyAging, Stanford Dogs, and Cars show how augmentations create temporal variations from one image. The first column shows originals (green); others apply augmentations by color: flip (red), zoom (blue), rotation (purple), color jitter (orange), shear (brown), translation (pink), blur (gray), and cutout (cyan), enriching the feature space with varied appearances.}
  \label{fig:vertical_time_aug3}
\end{figure*}

\begin{figure*}[tbp]
  \centering
  \begin{subfigure}[b]{0.35\textwidth}
    \centering
    \includegraphics[width=\textwidth,clip,trim=0 0 0 0]{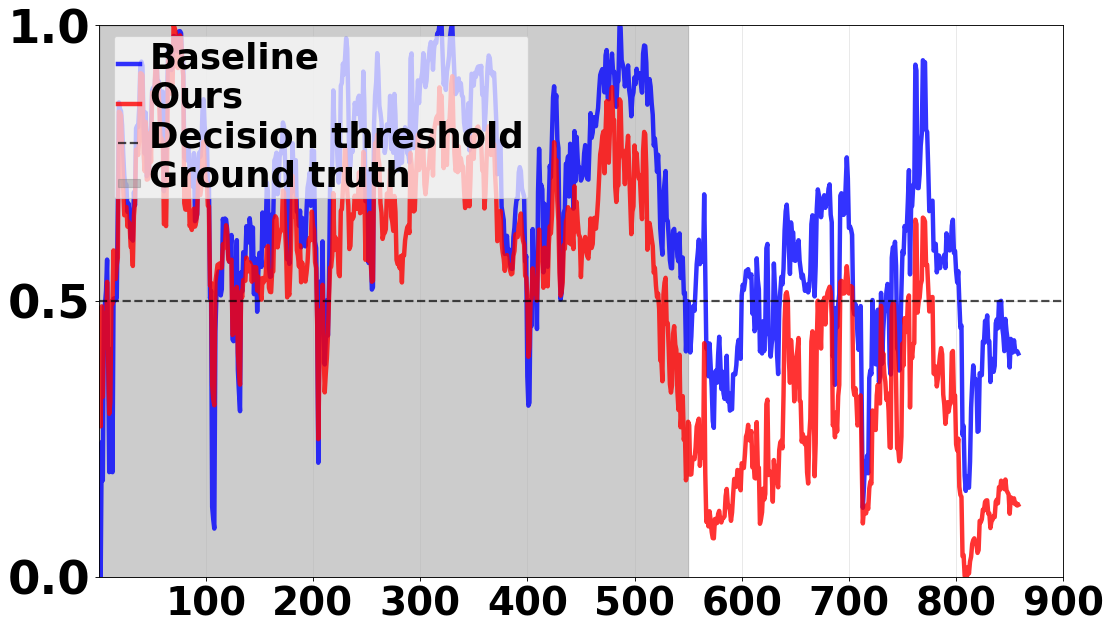}
    \caption{Assault}
    \label{fig:anomaly-a}
  \end{subfigure}%
  \hspace{5em}
  \begin{subfigure}[b]{0.35\textwidth}
    \centering
    \includegraphics[width=\textwidth,clip,trim=0 0 0 0]{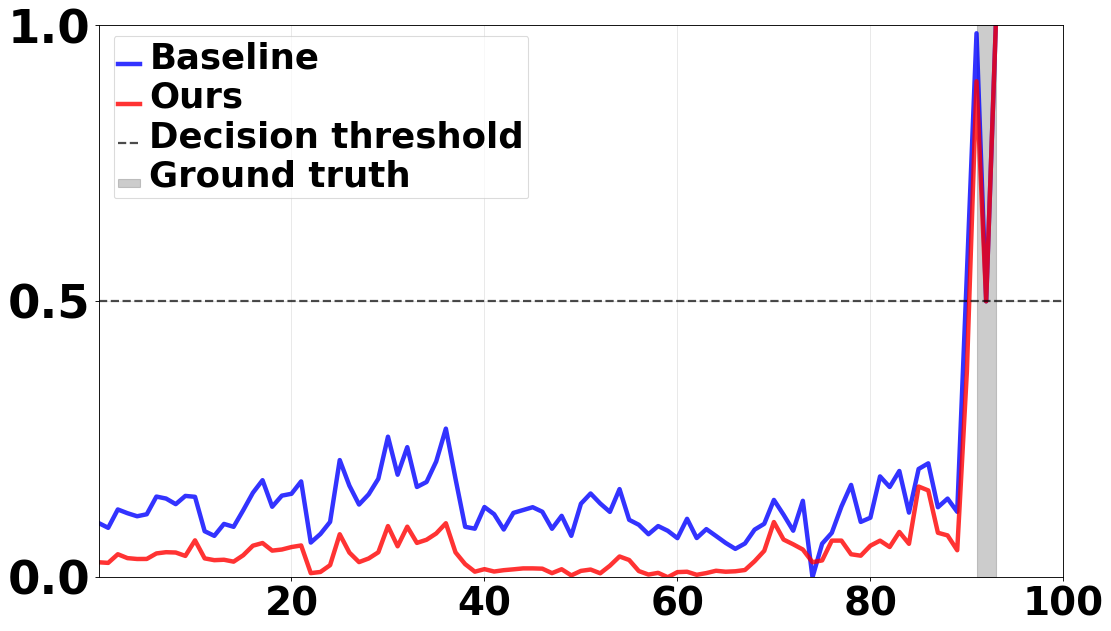}
    \caption{Explosion}
    \label{fig:anomaly-b}
  \end{subfigure}

  \vspace{1em}
  \begin{subfigure}[b]{0.35\textwidth}
    \centering
    \includegraphics[width=\textwidth,clip,trim=0 0 0 0]{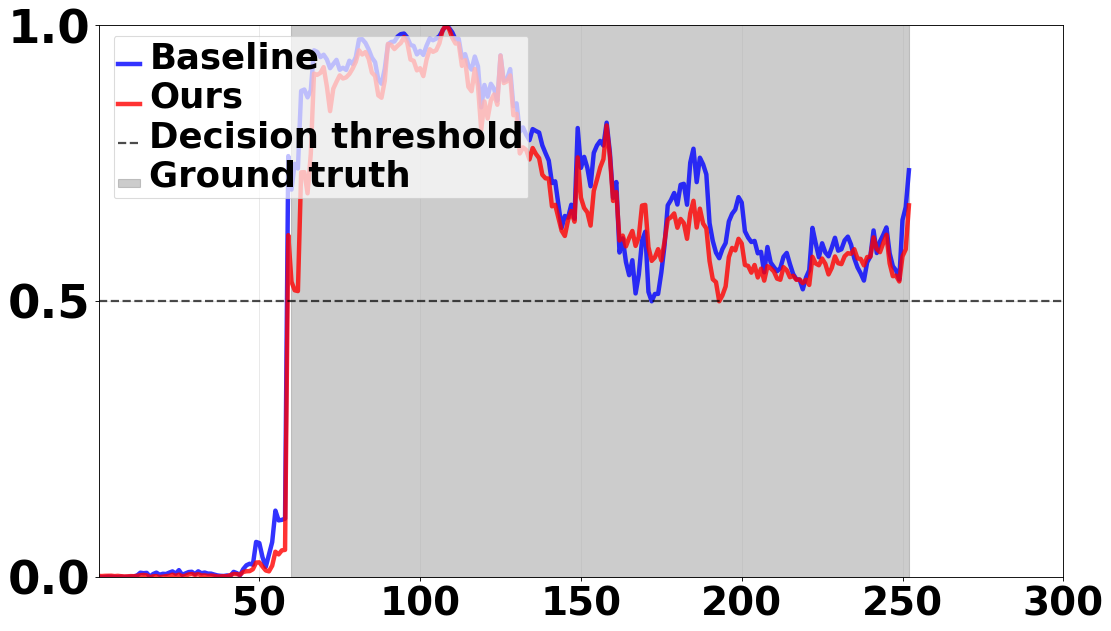}
    \caption{Explosion}
    \label{fig:anomaly-c}
  \end{subfigure}%
  \hspace{5em}
  \begin{subfigure}[b]{0.35\textwidth}
    \centering
    \includegraphics[width=\textwidth,clip,trim=0 0 0 0]{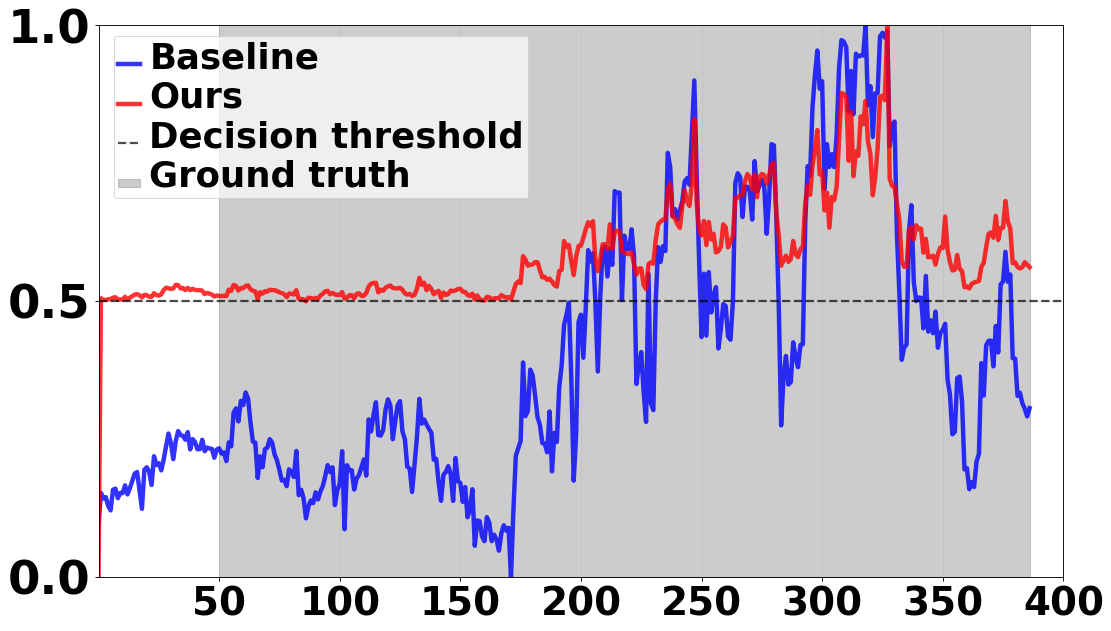}
    \caption{Traffic accident}
    \label{fig:anomaly-d}
  \end{subfigure}

  \vspace{1em}
  \begin{subfigure}[b]{0.35\textwidth}
    \centering
    \includegraphics[width=\textwidth,clip,trim=0 0 0 0]{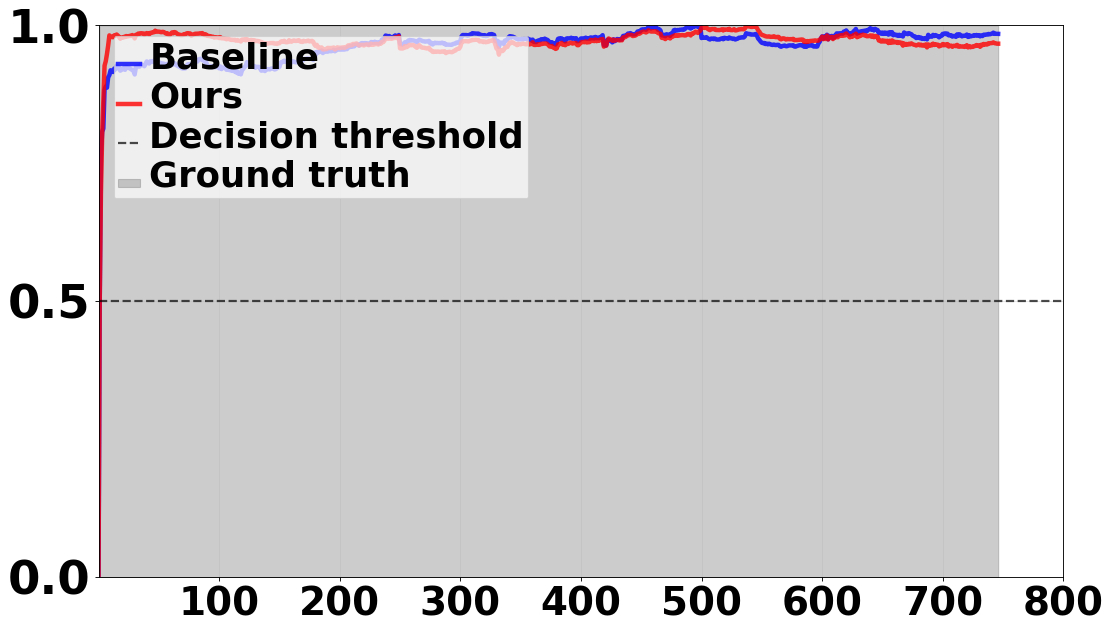}
    \caption{Fire}
    \label{fig:anomaly-e}
  \end{subfigure}%
  \hspace{5em}
  \begin{subfigure}[b]{0.35\textwidth}
    \centering
    \includegraphics[width=\textwidth,clip,trim=0 0 0 0]{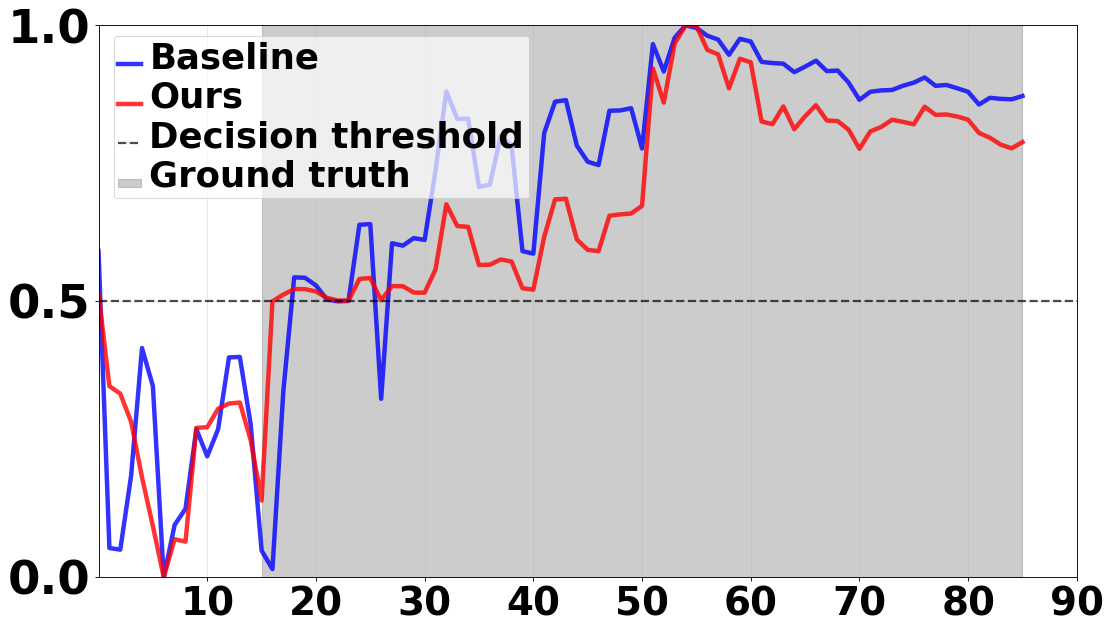}
    \caption{Object falling}
    \label{fig:anomaly-f}
  \end{subfigure}

  \vspace{1em}
  \begin{subfigure}[b]{0.35\textwidth}
    \centering
    \includegraphics[width=\textwidth,clip,trim=0 0 0 0]{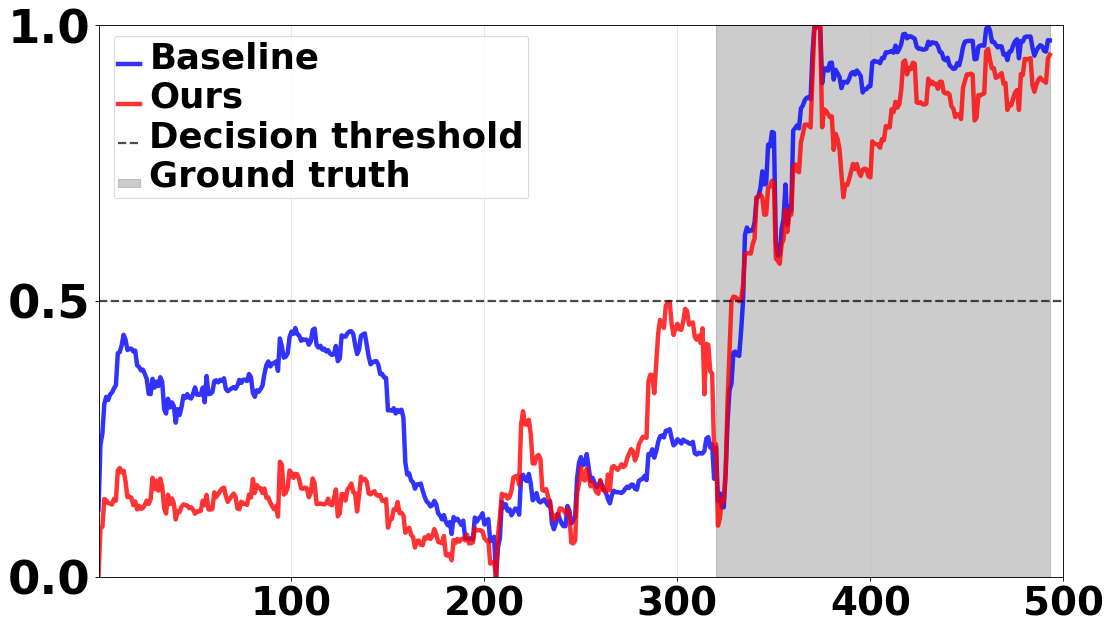}
    \caption{Fire}
    \label{fig:anomaly-g}
  \end{subfigure}%
  \hspace{5em}
  \begin{subfigure}[b]{0.35\textwidth}
    \centering
    \includegraphics[width=\textwidth,clip,trim=0 0 0 0]{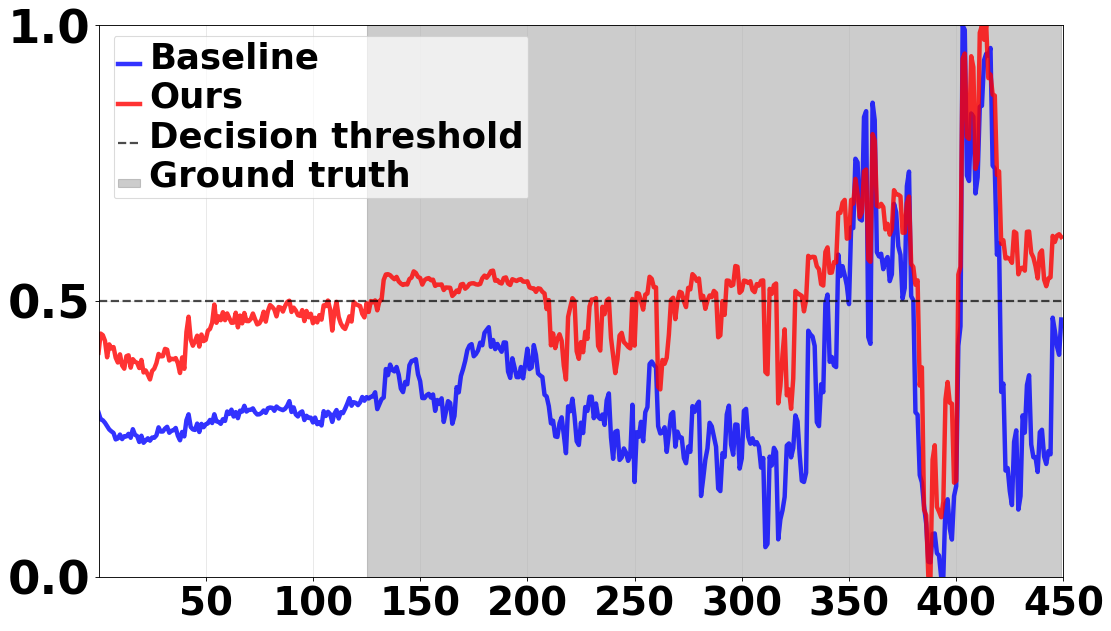}
    \caption{Traffic accident}
    \label{fig:anomaly-h}
  \end{subfigure}

  \caption{Anomaly prediction comparison. Grey regions indicate ground-truth anomalies. Blue and red curves show the baseline and our method. Our approach detects anomalies more accurately and earlier, with scores crossing the 0.5 threshold in closer alignment with the ground truth.}
  \label{fig:anomaly-testing-all}
\end{figure*}

\begin{figure*}[tbp]
  \centering
  \begin{subfigure}[b]{0.35\textwidth}
    \centering
    \includegraphics[width=\textwidth,clip,trim=0 0 0 0]{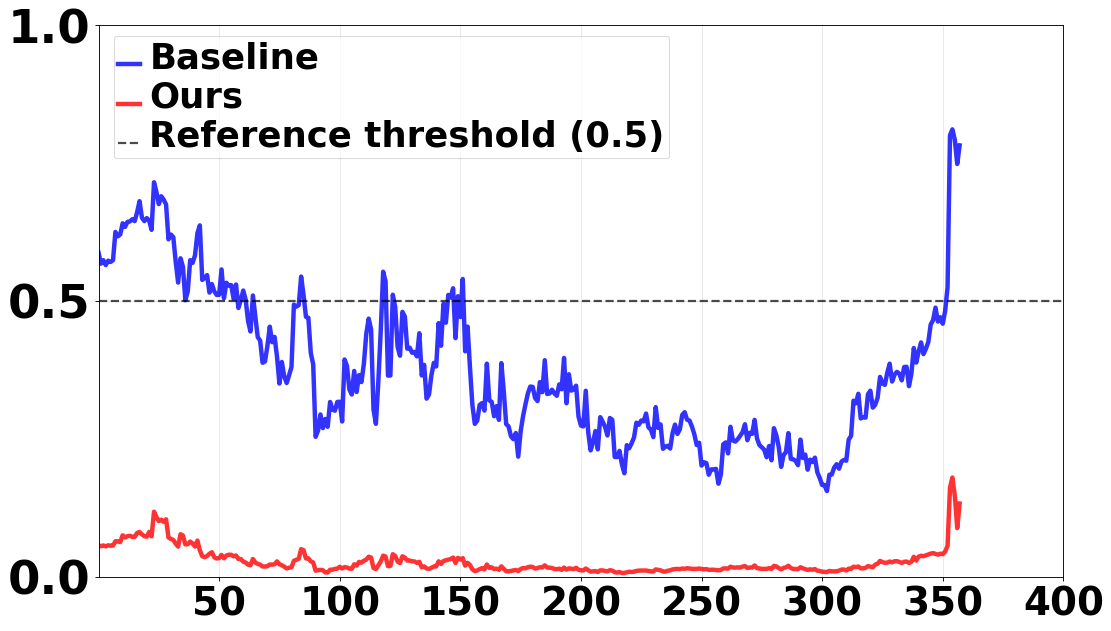}
    \caption{Road}
    \label{fig:normal-a}
  \end{subfigure}%
  \hspace{5em}
  \begin{subfigure}[b]{0.35\textwidth}
    \centering
    \includegraphics[width=\textwidth,clip,trim=0 0 0 0]{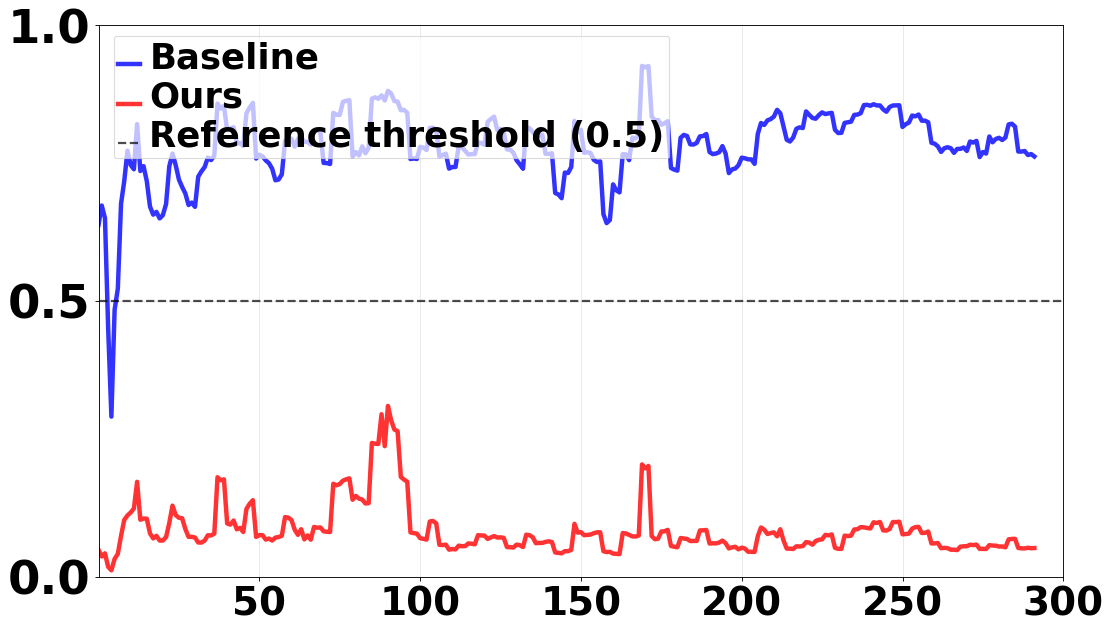}
    \caption{Train}
    \label{fig:normal-b}
  \end{subfigure}

  \vspace{1em}
  \begin{subfigure}[b]{0.35\textwidth}
    \centering
    \includegraphics[width=\textwidth,clip,trim=0 0 0 0]{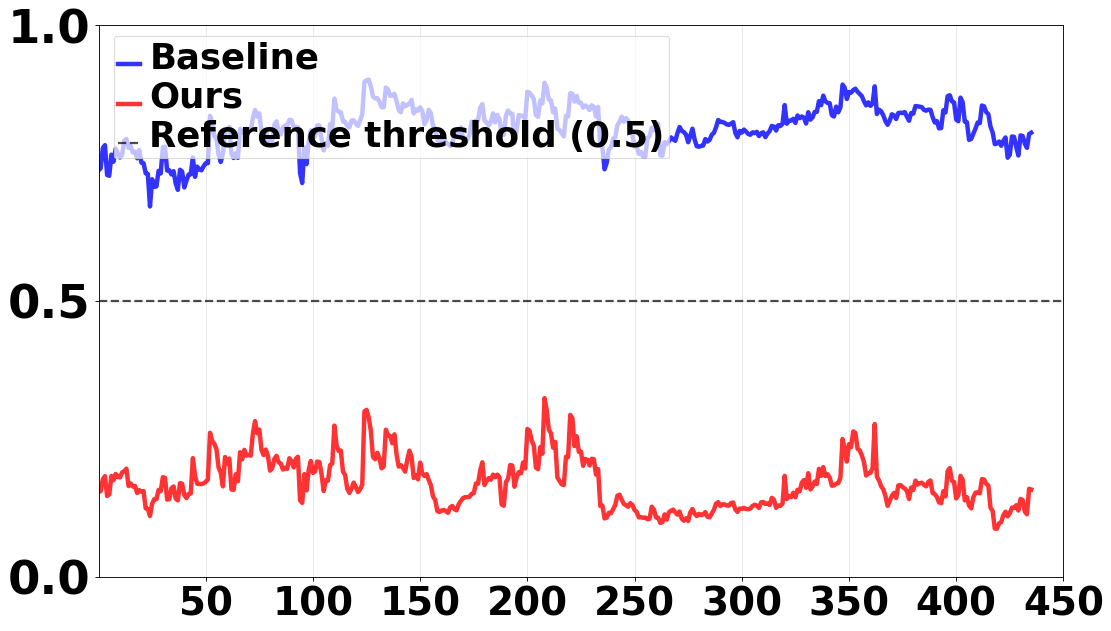}
    \caption{Street highview}
    \label{fig:normal-c}
  \end{subfigure}%
  \hspace{5em}
  \begin{subfigure}[b]{0.35\textwidth}
    \centering
    \includegraphics[width=\textwidth,clip,trim=0 0 0 0]{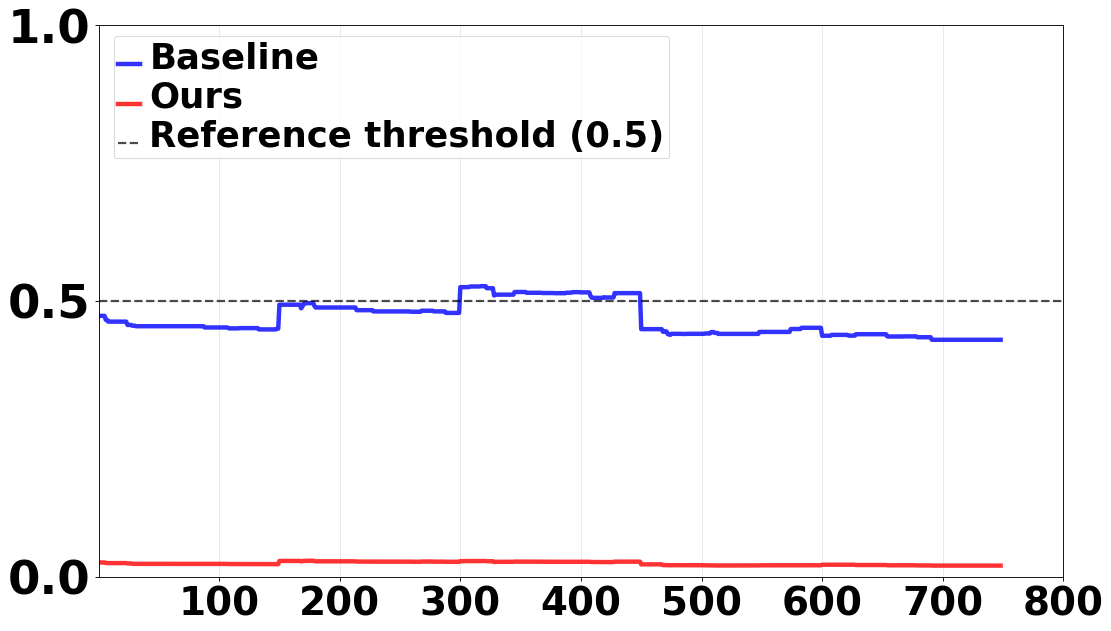}
    \caption{Sidewalk}
    \label{fig:normal-d}
  \end{subfigure}

  \vspace{1em}
  \begin{subfigure}[b]{0.35\textwidth}
    \centering
    \includegraphics[width=\textwidth,clip,trim=0 0 0 0]{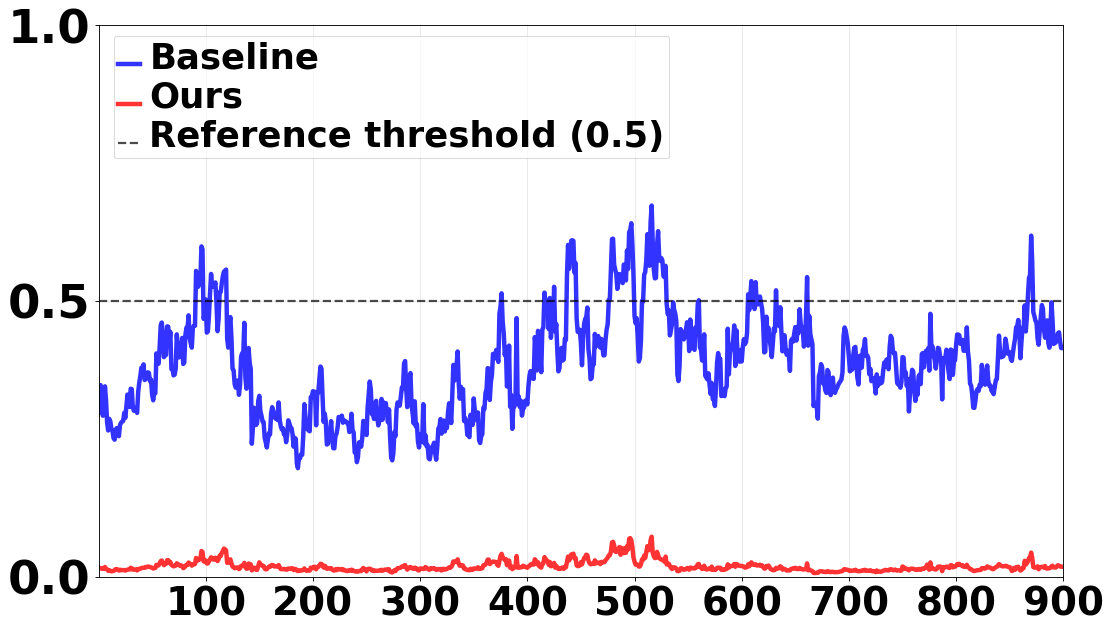}
    \caption{Train}
    \label{fig:normal-e}
  \end{subfigure}%
  \hspace{5em}
  \begin{subfigure}[b]{0.35\textwidth}
    \centering
    \includegraphics[width=\textwidth,clip,trim=0 0 0 0]{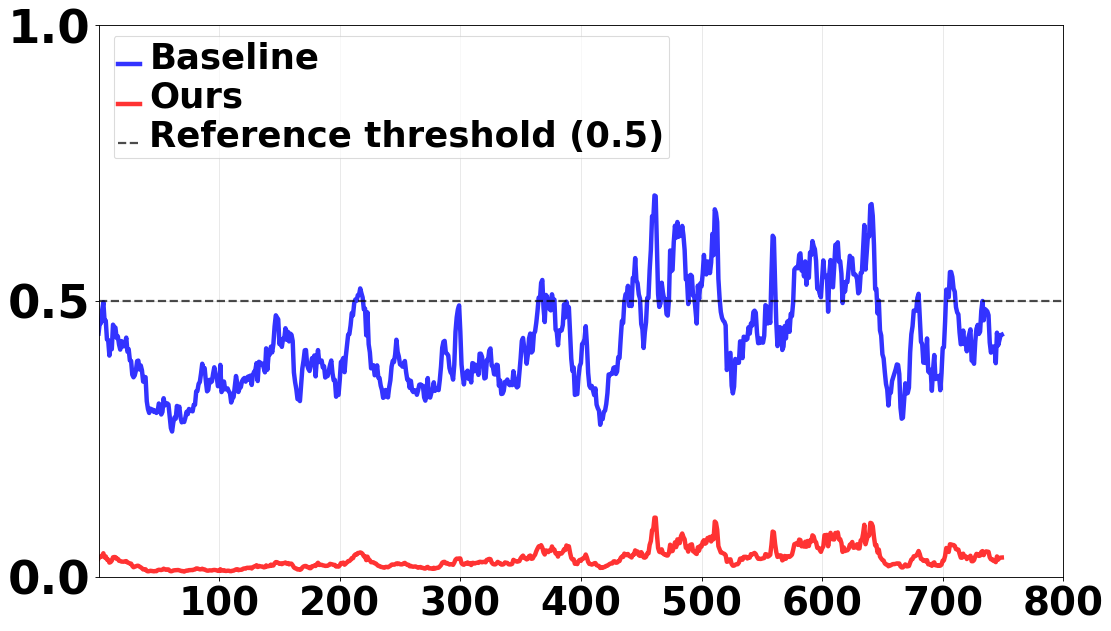}
    \caption{Sidewalk}
    \label{fig:normal-f}
  \end{subfigure}

  \vspace{1em}
  \begin{subfigure}[b]{0.35\textwidth}
    \centering
    \includegraphics[width=\textwidth,clip,trim=0 0 0 0]{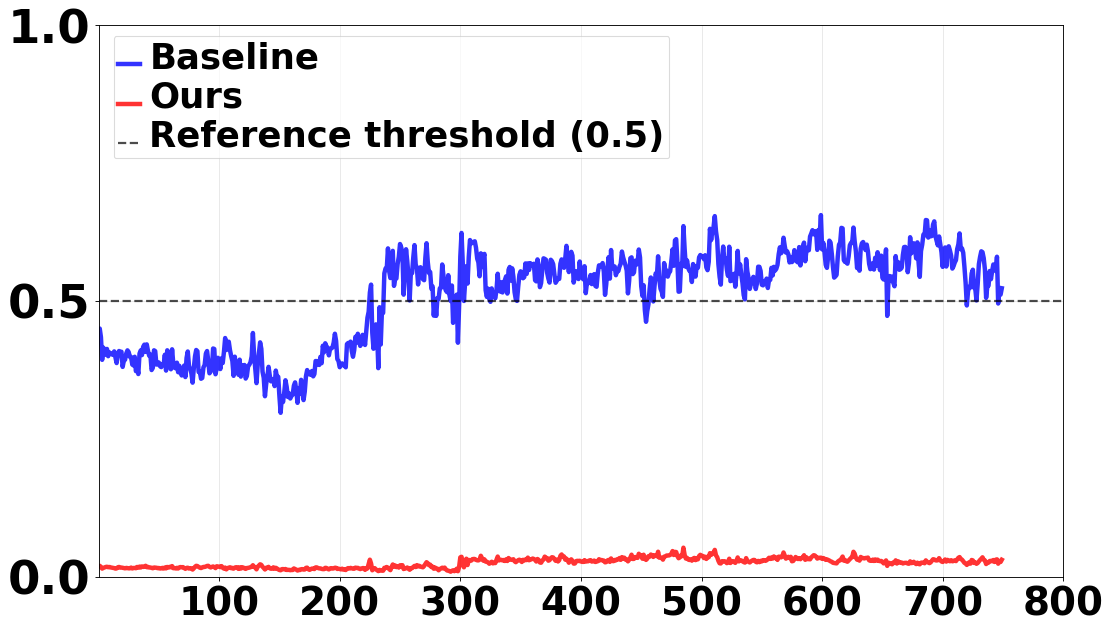}
    \caption{Road}
    \label{fig:normal-g}
  \end{subfigure}%
  \hspace{5em}
  \begin{subfigure}[b]{0.35\textwidth}
    \centering
    \includegraphics[width=\textwidth,clip,trim=0 0 0 0]{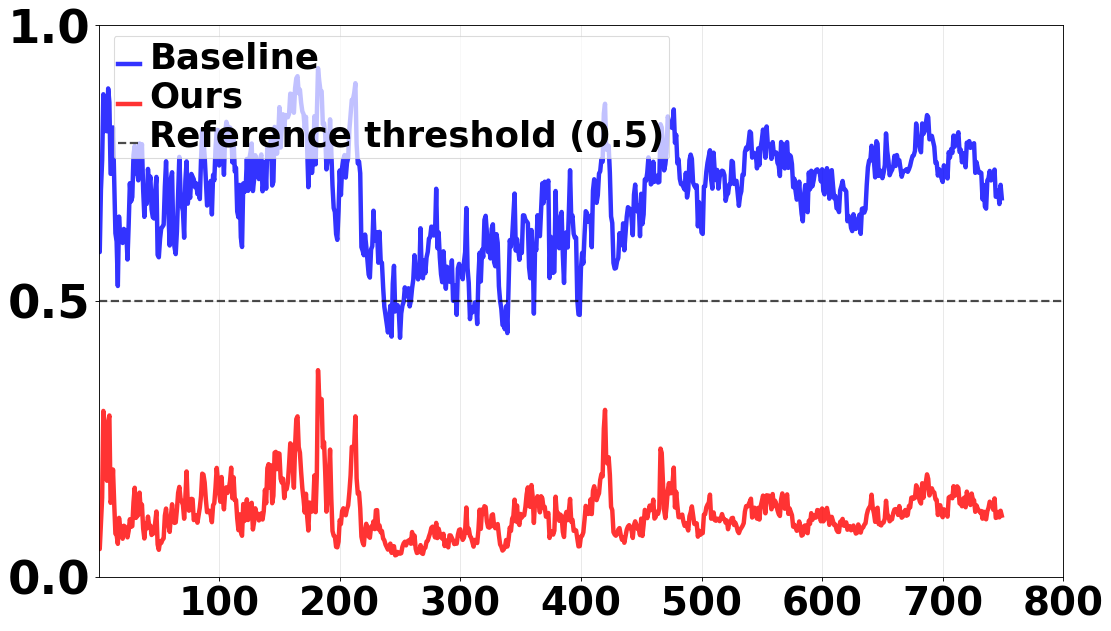}
    \caption{Sidewalk}
    \label{fig:normal-h}
  \end{subfigure}

  \caption{Anomaly prediction comparison. In these normal scenarios, our methods correctly avoid false positives, while the baseline method incorrectly flags them as anomalies.}
  \label{fig:normal-testing-all}
\end{figure*}

\subsection{E. Additional Visualizations}

\subsubsection{E.1. Additional Smooth Temporal Augmentations}

Below (Fig. \ref{fig:vertical_time_aug3}), we provide additional visualizations showcasing our smooth temporal augmentations applied to fine-grained and ultra-fine-grained image datasets, including Flowers-102, SoyAging, Stanford Dogs, and Stanford Cars. Each example starts with the original image (green, first column), followed by a sequence of augmented variants: horizontal flip (red), zoom (blue), rotation (purple), color jitter (orange), shear (brown), translation (pink), blur (gray), and cutout (cyan). These augmentations introduce realistic temporal variations from a single static image, enriching the feature space with diverse yet semantically consistent appearances, crucial for modeling temporal dynamics in the absence of natural video sequences.

\subsubsection{E.2. Additional Video Anomaly Detection Visualizations}

Below (Fig. \ref{fig:anomaly-testing-all} and \ref{fig:normal-testing-all}), we present additional visualizations of video anomaly detection results on the MSAD dataset. These examples demonstrate that our method detects anomalies not only with greater accuracy but also at earlier time steps compared to competing approaches, highlighting its effectiveness in capturing subtle temporal deviations.

\subsection{F. Additional Discussions}

This section provides further explanation of several aspects of SEQ, including its 
generalization behaviour, computational characteristics, temporal augmentation design, 
and the role of backbone choices and ablations.

\textbf{Generalizability.} SEQ models temporal structure by learning class-level temporal distributions rather than 
memorizing fixed temporal patterns. During training, each episode samples diverse 
support sequences whose augmentation parameters evolve smoothly over time. The resulting 
exemplars act as barycenters of multiple trajectories, capturing characteristic 
prediction-space evolution for each class. This episodic diversity encourages the model 
to form robust temporal prototypes that generalize across input variations. Moreover, the 
FC+Softmax mapping preserves trajectory smoothness, allowing temporal continuity to be 
maintained throughout the prediction sequence.

\textbf{Complexity.} Although Soft-DTW-based alignment is used during training, it operates on 
$\tau \times \tau'$ matrices of class probability vectors and thus has moderate 
computational cost. In the image domains considered in this work, $\tau \le 5$ and the 
support sets are small, making alignment efficient and lightweight. Importantly, SEQ 
introduces no additional cost at inference time: the classifier reduces to a single 
fully connected layer identical to the static baseline, with no need for alignment or 
temporal matching. This preserves the architectural simplicity that motivates the 
framework and differentiates it from recurrent or transformer-based temporal models.

\textbf{Augmentation.} The temporal augmentation strategy is designed to impose a temporal continuity bias 
rather than simulate physically accurate motion. Augmentation parameters such as rotation, 
translation, or color intensity evolve linearly across virtual timesteps, ensuring that 
the induced variation is smooth and coherent. Applying identical augmentation schedules 
to both support and query sequences prevents artificial discrepancies and ensures that 
the learning signal is governed by feature-level temporal evolution rather than 
augmentation artifacts. This provides a controlled setting in which the classifier can 
learn trajectory consistency while remaining agnostic to the exact nature of image-level 
transformations. Extensions based on generative temporal augmentations or video test-time 
adaptation represent promising avenues for future work.

\textbf{Backbone fairness.} All comparisons within each dataset use the same frozen backbone for both the baseline 
and the temporal variants. This ensures that improvements arise solely from modeling 
temporal structure rather than from differences in representation quality. Different 
datasets use different backbones only to match established conventions in prior work 
and dataset-specific domain characteristics (\eg, CLIP-ViT for Cars, ViT-B/16 for Dogs, CLE-ViT (Swin-B/448, IN-21K) for ultra-fine SoyAging). Experiments using a unified backbone across 
Cars and Dogs demonstrate that the relative gains of SEQ remain stable, indicating 
backbone-invariance of the temporal modeling benefits.

\textbf{Ablations \& statistics.} Ablation studies emphasize the complementary contributions of the different loss components. Removing any single loss term leads to a decrease in accuracy. For example, on the Flowers-102 dataset, we observe the following results: Baseline $97.5 \pm 0.5$, without $\mathcal{L}_{\text{align}}$ $97.6 \pm 0.4$, without $\mathcal{L}_{\text{smooth}}$ $98.0 \pm 0.3$, without $\mathcal{L}_{\text{CE}}$ $96.8 \pm 0.6$, and the full SEQ objective $98.4 \pm 0.3$ ($p<0.01$ across tasks). 

Specifically, removing alignment, smoothness, or semantic supervision consistently degrades performance across datasets. The complete objective, which integrates temporal prototype alignment, semantic consistency, and smooth temporal evolution, produces the most coherent prediction trajectories and achieves the highest recognition accuracy, highlighting the importance of combining all components.


\end{document}